\documentclass[11pt]{article}
\usepackage{listings}
\lstdefinestyle{pseudocode}{language=Python, basicstyle=\ttfamily\small, keywordstyle=\color{blue}, commentstyle=\color{gray}, stringstyle=\color{red}, showstringspaces=false, breaklines=true}
\usepackage{microtype}
\usepackage{graphicx}
\usepackage{booktabs}
\usepackage{amsmath}
\usepackage{amsfonts}

\usepackage{amssymb}
\usepackage{bm}
\usepackage{enumitem}
\usepackage{xspace}
\usepackage{makecell}
\usepackage{multirow}
\usepackage{algorithm}
\usepackage{algpseudocode}
\usepackage{arxiv}
\usepackage{hyperref}
\usepackage{subcaption}

\newcommand{\methodname}{HLA-WM}
\usepackage{url}
\usepackage{caption}
\newcommand{\reporttitle}{HLA-WM: Hybrid Linear Attention for Long-Horizon Video World Models}
\newcommand{\reportshorttitle}{Short Title}
\newcommand{\reportauthors}{
\textbf{Zhuokun Chen}\textsuperscript{\normalfont 1}  \quad
\textbf{Feng Chen}\textsuperscript{\normalfont 2} \quad
\textbf{Xi Lin}\textsuperscript{\normalfont 3} \quad
\textbf{Xiyu Wu}\textsuperscript{\normalfont 3} \quad
\textbf{Jiahao He}\textsuperscript{\normalfont 3}  \\
\textbf{Jianfei Cai}\textsuperscript{1} \quad
\textbf{Bohan Zhuang}\textsuperscript{3}
}
\newcommand{\reportaffiliations}{\textsuperscript{1}Monash University \quad
\textsuperscript{2}The University of Adelaide \quad
\textsuperscript{3}Zhejiang University}

\newcommand{\reportproject}{\href{https://caesarhhh.github.io/hla-wm/}{https://caesarhhh.github.io/hla-wm/}}
\newcommand{\reportemail}{\href{mailto:bohan.zhuang@zju.edu.cn}{bohan.zhuang@zju.edu.cn}}
\newcommand{\reportorcid}{\href{https://orcid.org/0000-0000-0000-0000}{First Author}}
\newcommand{\reportkeywords}{Video World Models, Linear Attention, Long-Range Memory, Gated DeltaNet}

\newcommand{\reportpdfauthors}{First Author, Second Author, Third Author}
\newcommand{\reportlogoheight}{0.62cm}

\techreportlabel{ZIP Lab Technical Report}
\techreportshorttitle{\reportshorttitle}
\techreportsetlogoheight{\reportlogoheight}
\techreportproject{\reportproject}
\techreportemail{\reportemail}
\techreportorcid{\reportorcid}
\techreportkeywords{\reportkeywords}

\techreportlogos{%
  \techreportlogo{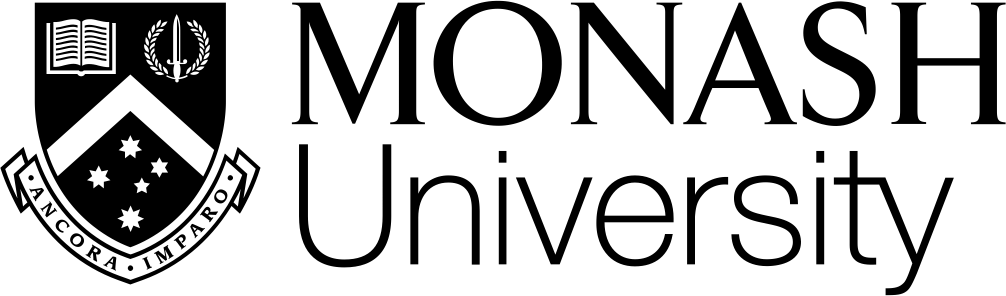}
  \techreportlogo{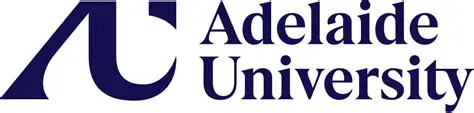}
  \techreportlogo{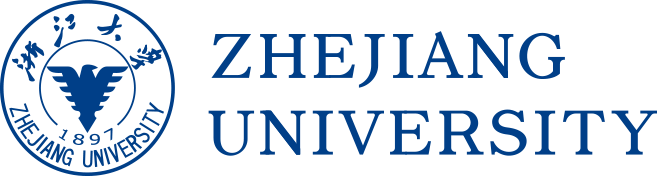}
}

\hypersetup{
  colorlinks=true,
  linkcolor=ziplabblue,
  citecolor=ziplabblue,
  urlcolor=ziplabblue,
  pdftitle={\reporttitle},
  pdfauthor={\reportpdfauthors},
  pdfsubject={ZIP Lab technical report},
  pdfkeywords={\reportkeywords}
}

\begin{document}

\ziplabmaketitle{\reporttitle}{\reportauthors}{\reportaffiliations}{%
  \begin{center}
  \end{center}
}
\ziplabendtitle{Long-horizon video world models require persistent memory to preserve scene consistency over extended rollouts.\
Softmax attention retains the full generation history through a growing KV cache, whereas recurrent linear attention compresses history into fixed-size states with substantially lower memory cost.\
However, we identify severe long-range forgetting in Gated DeltaNet (GDN), where information from distant but relevant scenes is progressively attenuated by subsequent state updates.\
To address this limitation, we propose HLA-WM, a training-free \textit{hybrid linear-attention framework} that combines coarse-grained geometry-guided retrieval with fine-grained recurrent linear-state computation.\
HLA-WM exploits the affine structure of GDN to cache compact chunk-wise transition summaries, retrieve scene-relevant historical chunks using camera geometry, and recompose them into query-specific recurrent states.\
On the $60$-second SANA-WM-Bench, HLA-WM improves all six aggregate revisit-consistency and camera-control metrics of the base autoregressive generator without additional training, including a $0.74$ dB PSNR gain and a $28.5\%$ reduction in rotation error.\
The improvements persist after downstream refinement and generalize to MBench-A, where HLA-WM consistently improves all three revisit-consistency metrics across all four subsets and all evaluated inference modes over $547$ samples.\
At a $60$-second context, HLA-WM reduces historical-state memory by $12\times$ relative to full KV caching while incurring at most a $1.6\%$ reduction in inference throughput.\
These results demonstrate that selectively addressable recurrent memory can improve long-range scene recall while preserving the efficiency advantages of GDN.}

\section{Introduction}

World models predict environment dynamics, enabling agents to anticipate future observations and reason about the consequences of their actions~\citep{ha2018world,hafner2025mastering}.
Recent video world models realize such simulation through autoregressive generation conditioned on agent actions or camera trajectories~\citep{zhang2025matrix,he2025matrix,mao2025yume,zhu2026astra,zhu2026sana}.
Beyond short-term temporal continuity, long-horizon simulation requires previously observed scenes to remain recoverable after they leave the recent context~\citep{wu2026video,hong2025relic,wang2026matrix}.
When the camera later returns, the model should recover the earlier scene structure rather than generate a semantically similar but spatially inconsistent view.
Full-history key-value (KV) caching provides direct access to past observations, but its storage cost grows continuously with the rollout length~\citep{yang2026longlive,yesiltepe2026videomla}.
Recurrent linear attention offers a compact alternative by compressing history into fixed-size states~\citep{katharopoulos2020transformers,li2026attend,zhu2026sana}.
This raises a central question for long-horizon world modeling: \textit{how can recurrent models retain efficient memory while still recovering distant scene information when it becomes relevant again?}

\begin{figure}[t]
    \centering
    \includegraphics[width=\linewidth]
    {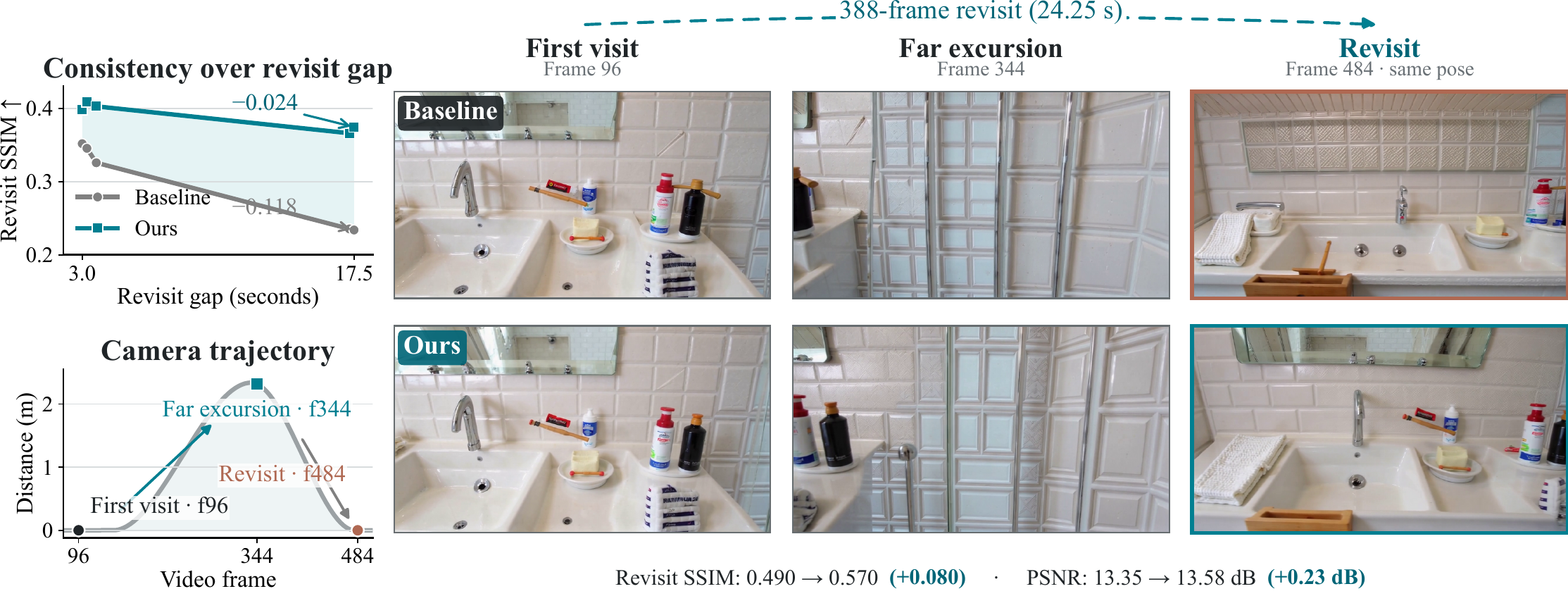}
    \caption{
    \textbf{Long-range forgetting in GDN-based world-model rollouts.}
    \textbf{Upper left:} Revisit SSIM over increasing temporal gaps.
    \textbf{Lower left:} Camera displacement along the qualitative trajectory.
    \textbf{Right:} Qualitative comparison of the first visit, far excursion, and revisit at the similar camera pose. After the long excursion, the baseline SANA-WM shows noticeable changes in scene layout and object arrangement, whereas our \methodname~better preserves the first-visit scene content.
    }
    \label{fig:intro_forgetting}
\end{figure}

Among modern recurrent linear-attention architectures, Gated DeltaNet (GDN) combines data-dependent decay with delta-rule updates to manage information within a compact recurrent state~\citep{yang2025gated}.
Nevertheless, we observe a clear \textbf{long-range forgetting} problem in GDN-based video world models.
As illustrated in Fig.~\ref{fig:intro_forgetting}, SANA-WM exhibits a pronounced decline in revisit consistency as the temporal gap increases, while the qualitative example shows substantial changes in scene structure after the camera leaves and later returns.
Such revisits are particularly challenging because information that is no longer locally relevant may become essential again after a long sequence of intermediate observations.
This behavior suggests that compact recurrent memory alone is insufficient to reliably preserve distant scene information over extended rollouts.

To understand this limitation, we analyze the recurrent state transition in GDN.
Since the full history is compressed into a single recurrent state, earlier information must survive all intervening updates to remain recoverable, motivating a memory organization in which compact historical states remain independently addressable.

We therefore ask whether a pretrained recurrent video world model can recover distant scene history without retaining a full token-level KV cache or requiring additional training.
To this end, we propose \textbf{\methodname}, a training-free \textit{hybrid linear-attention framework} that combines coarse-grained geometry-guided retrieval with fine-grained recurrent linear-attention access.
\methodname~exploits the affine structure of GDN to cache compact chunk-wise summaries of historical state transitions, selectively retrieves relevant summaries using camera geometry, and recomposes them into a query-specific recurrent state.
This establishes an intermediate memory regime between a single recurrent state and token-level full-history caching, enabling selective access to distant history while retaining compact linear-state storage.

For camera-controlled world-model rollouts, camera metadata provides a natural addressing signal without requiring a learned router~\citep{yu2025context,xiao2026worldmem}.
\methodname~uses a lightweight camera-frustum overlap proxy derived from poses, intrinsics, and a scene-level depth estimate to identify relevant historical chunks.
The selected summaries, together with the first chunk as a persistent sink and recent context for local continuity, are recomposed as retrieved context in chronological order without replaying historical visual tokens.
Selective recomposition is applied only to the main GDN memory, while camera-control and local convolutional states retain their original recurrent updates.
All pretrained parameters remain unchanged.
Thus, \methodname~changes how historical memory is stored and accessed rather than retraining the underlying generator.

Experiments on SANA-WM-Bench~\citep{zhu2026sana} and MBench-A~\citep{zhang2026mbench} demonstrate consistent improvements in long-range revisit fidelity.
At Stage~1, where \methodname~directly modifies recurrent memory, it also improves camera controllability by reducing accumulated structural drift and maintaining a more stable spatial reference during generation.
Downstream refinement introduces mode-dependent trade-offs, while the long-range consistency gains remain evident across both benchmarks.
On MBench-A, all three revisit-consistency metrics improve across all four subsets and all evaluated inference modes.
At a $60$-second context, \methodname~requires $12\times$ less historical-state storage than full KV caching while incurring at most a $1.6\%$ reduction in inference throughput across the evaluated pipelines.
These results show that selectively addressable recurrent memory can substantially improve long-range scene recall.

Our main contributions are summarized as follows:
\begin{itemize}[leftmargin=*]

\item We empirically characterize long-range forgetting in GDN-based video world models and show how accumulated recurrent transitions progressively weaken access to distant scene information.

\item We propose \methodname, a training-free hybrid linear-attention framework that equips recurrent video world models with selectively addressable historical memory through compact affine summaries, camera-guided addressing, and chronological state recomposition.

\item Extensive experiments on SANA-WM-Bench and MBench-A demonstrate improved long-range scene consistency.
On the $60$-second SANA-WM-Bench, \methodname~improves the base autoregressive generator by $0.74$ dB in PSNR and reduces rotation error by $28.5\%$ without training.

\end{itemize}

\section{Related Work}

\textbf{Video World Models.}
Recent advances in video generation have enabled world models to predict future visual observations conditioned on actions, camera trajectories, or other control signals, supporting interactive and long-horizon environment simulation~\citep{zhang2025matrix,he2025matrix,mao2025yume,zhu2026astra,zhu2026sana}. To maintain scene consistency over extended rollouts, existing approaches preserve historical information in different forms. Frame-based methods retain selected historical observations as references for future generation~\citep{wu2026video,wang2026matrix}. Token- or KV-based methods preserve historical visual tokens or attention caches to provide direct access to past context~\citep{hong2025relic,samuel2026fast,wu2026addressable}. Feature-memory methods summarize past observations into compact latent representations or dedicated memory tokens~\citep{wei2026geometry}. Spatial-memory approaches instead maintain explicit geometric or 3D representations that can be queried according to the current viewpoint~\citep{wu2026video,wang2026matrix,team2026alayaworld}. Although these mechanisms improve long-range consistency, they introduce additional storage and retrieval overhead that can grow with the rollout. In particular, frame-based memories may require retrieved observations to be re-encoded and their attention caches recomputed, while token, feature, and spatial memories require maintaining and querying additional historical representations, limiting their efficiency for long-horizon generation.

\textbf{Linear Attention for World Models.}
Linear attention reformulates attention through associative computation, allowing historical information to be summarized into fixed-size recurrent states instead of maintaining a KV cache that grows with sequence length~\citep{katharopoulos2020transformers,schlag2021linear}. Building on this formulation, subsequent works improve the expressiveness and memory management of linear states through more sophisticated recurrent updates. Gated Linear Attention introduces data-dependent decay to selectively forget historical information~\citep{pmlr-v235-yang24ab}, while DeltaNet employs the delta rule to enable targeted updates to the recurrent memory~\citep{yang2024parallelizing}. Gated DeltaNet (GDN) further combines these two mechanisms, providing both adaptive forgetting and selective memory updates within a fixed-size state~\citep{yang2025gated}. More recently, recurrent linear attention has been extended to autoregressive video generation, where linear states serve as compact cross-frame memory for long-form generation~\citep{li2026attend,zhu2026sana}. Despite their favorable memory efficiency, these methods continuously compress historical observations into a limited recurrent state, making distant information increasingly susceptible to attenuation and interference from subsequent updates during long-horizon generation.

\begin{figure}[t]
\centering
\includegraphics[width=\linewidth]{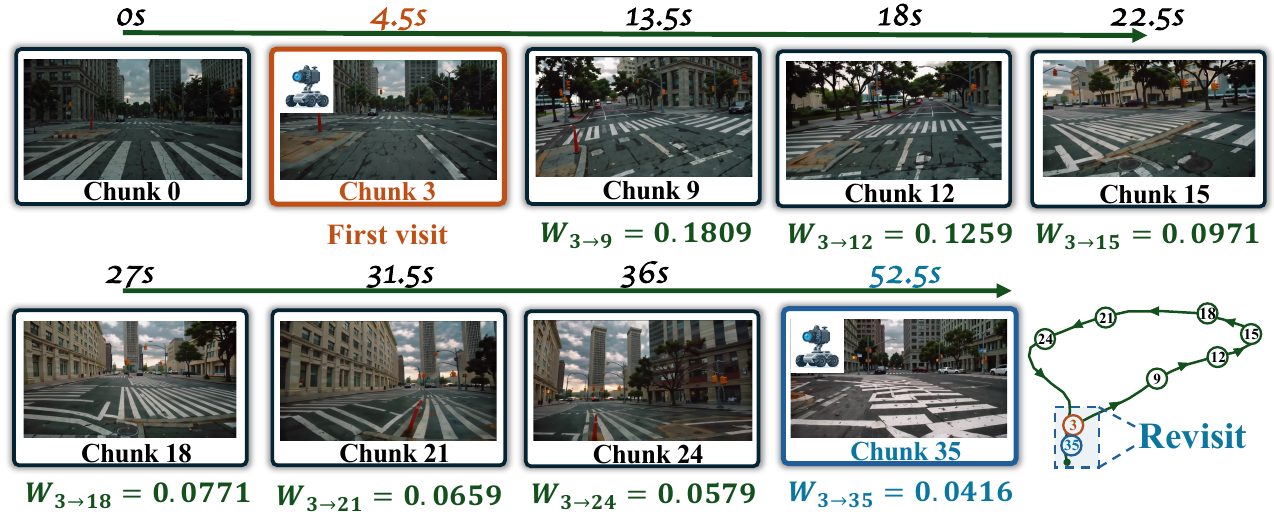}
\caption{
\textbf{Long-range forgetting in GDN.} The camera follows a long loop and eventually returns near the region observed at Chunk $3$ around Chunk $35$. We visualize representative generated chunks along the trajectory together with the cumulative retention $W_{3\rightarrow j}$ of Chunk $3$. Although the early scene becomes relevant again near the end of the rollout, its retained influence continuously decreases as intermediate chunks are generated, reaching only $0.0416$ at Chunk $35$.
}
\label{fig:gdn_forgetting}
\end{figure}

\section{Empirical Insights}
\label{sec:empirical_insights}

We first investigate the long-range forgetting behavior of GDN in SANA-WM~\citep{zhu2026sana}. At the chunk level, the recurrent update of chunk $i$ can be written as
\begin{equation}
S_i = S_{i-1} A_i + B_i,
\label{eq:gdn_chunk_affine}
\end{equation}
where $A_i$ denotes the state transition applied to historical memory and $B_i$ represents the information newly written by the current chunk. Expanding the recurrence over multiple chunks gives
\begin{equation}
S_j
=
\cdots
+
B_i A_{i+1} A_{i+2} \cdots A_j
+
\cdots
+
B_j.
\label{eq:gdn_chunk_composition}
\end{equation}
Thus, information written by an early chunk is repeatedly transformed by all subsequent transition matrices before it can influence a distant future scene. We quantify its cumulative retention from chunk $i$ to chunk $j$ using the average norm of the intervening transition product:
\begin{equation}
W_{i\rightarrow j}
=
\operatorname{AvgNorm}
\left(
A_{i+1} A_{i+2} \cdots A_j
\right),
\label{eq:gdn_retention}
\end{equation}
where a smaller $W_{i\rightarrow j}$ indicates stronger attenuation from chunk $i$ to $j$.

Figure~\ref{fig:gdn_forgetting} illustrates this phenomenon on a representative trajectory from the SANA-WM Benchmark. The camera leaves the region observed at Chunk $3$, traverses a large loop, and eventually returns to a nearby viewpoint at Chunk $35$. Since the two chunks observe highly overlapping regions, the information stored at Chunk $3$ should remain particularly relevant when generating Chunk $35$. However, its retained influence steadily diminishes as the rollout progresses and drops to only $0.0416$ by the time the camera revisits the scene. Correspondingly, the scene generated at Chunk $35$ deviates substantially from the earlier observation, indicating that the model can no longer effectively recover the distant but relevant scene information.

More importantly, most intermediate chunks correspond to different viewpoints and are only \textbf{weakly related} to the scene revisited at Chunk $35$. Nevertheless, their state-transition matrices are still sequentially applied to the memory written by Chunk $3$, causing that memory to undergo repeated transformations even when the intermediate observations provide little useful information for the later revisit. As a result, scene information that becomes relevant again can be substantially attenuated before it is needed. This exposes a fundamental mismatch in conventional GDN memory: \textit{historical information is updated according to temporal progression, whereas its relevance in world-model rollouts is often determined by spatial and geometric proximity}. This mismatch motivates selectively preserving and composing scene-relevant historical states instead of propagating all past information uniformly through every intermediate chunk.

\section{Method}
\label{sec:method}

\subsection{Overview}

\begin{figure}[t]
\centering
\includegraphics[width=\linewidth]{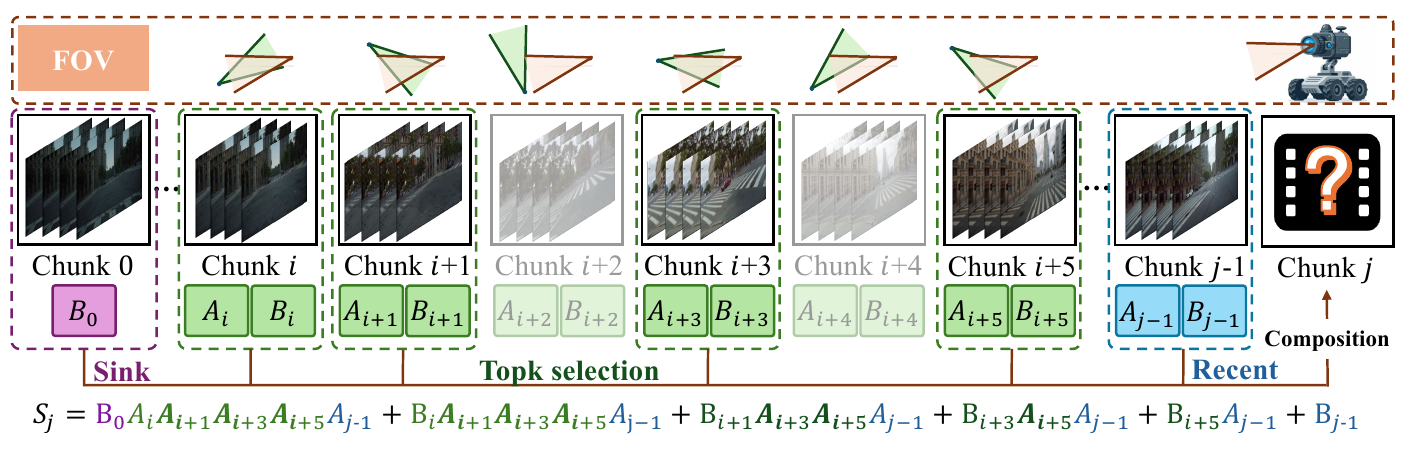}
\caption{
Overview of \methodname. Historical video chunks are represented by independent GDN summaries. Given the current query, geometry-guided retrieval selects scene-relevant historical chunks together with persistent sink and recent context. The selected summaries are then composed in chronological order to construct a query-specific recurrent state.
}
\label{fig:method_overview}
\end{figure}

We propose \methodname, a hybrid linear-attention framework for long-horizon video world models. The key idea is to replace the single accumulated GDN memory with a collection of independently addressable chunk-wise memories. Specifically, we partition the generated sequence into temporal chunks and summarize the GDN transition induced by each chunk. For each query chunk, \methodname~uses camera geometry to retrieve a sparse set of scene-relevant historical chunks and selectively composes their summaries into the recurrent state used for generation. As illustrated in Fig.~\ref{fig:method_overview}, this design combines coarse-grained geometry-guided memory retrieval with recurrent linear attention, allowing distant but relevant information to bypass unrelated intermediate updates while preserving the compactness of linear-state memory.

\subsection{Chunk-wise GDN Memory}
\label{sec:chunk_memory}

Gated DeltaNet maintains a recurrent linear state through token-wise updates~\citep{yang2025gated}.
Let $t$ index tokens in the recurrent scan order. For token $t$, the state update is
\begin{equation}
S_t
=
\gamma_t S_{t-1}
+
\beta_t
\left(
v_t - S_{t-1}k_t
\right)k_t^{\top},
\label{eq:gdn_token_update}
\end{equation}
where $k_t$ and $v_t$ are the key and value vectors projected from the current token representation, $\gamma_t$ controls memory decay, and $\beta_t$ controls the strength of the delta update.
Rearranging Eq.~\ref{eq:gdn_token_update} gives
\begin{equation}
S_t = S_{t-1}A_t + B_t,
\label{eq:gdn_token_affine}
\end{equation}
where
\begin{equation}
A_t = \gamma_t I - \beta_t k_tk_t^{\top},
\qquad
B_t = \beta_t v_tk_t^{\top}.
\label{eq:gdn_token_ab}
\end{equation}

This affine form can be directly composed over consecutive tokens.
Consider chunk $i$ containing $L$ tokens, and let $(i,\ell)$ denote its $\ell$-th token.
Unrolling the recurrence within the chunk gives
\begin{align}
S_i
={}&
S_{i-1}A_{i,1}A_{i,2}\cdots A_{i,L}
\nonumber +
B_{i,1}A_{i,2}\cdots A_{i,L}
+\cdots+
B_{i,L}.
\label{eq:gdn_chunk_unroll}
\end{align}
Therefore, the entire chunk can be represented by a single affine transition
\begin{equation}
S_i = S_{i-1}A_i + B_i,
\label{eq:gdn_chunk_summary}
\end{equation}
where
\begin{equation}
A_i = A_{i,1}A_{i,2}\cdots A_{i,L},
\qquad
B_i =
\sum_{\ell=1}^{L}
B_{i,\ell}
\prod_{u=\ell+1}^{L}A_{i,u}.
\label{eq:gdn_chunk_ab}
\end{equation}
Here, the products follow the recurrent scan order, with an empty product defined as the identity.
The pair $(A_i,B_i)$ therefore provides a sufficient summary of chunk $i$: $A_i$ describes how the chunk transforms existing memory, while $B_i$ represents the information written by the chunk itself.

Equivalently, their summaries follow the composition rule
\begin{equation}
(A_i,B_i)\star(A_j,B_j)
=
(A_iA_j,B_iA_j+B_j).
\label{eq:summary_composition}
\end{equation}
The operation is associative, enabling an arbitrary subset of historical chunk summaries to be efficiently recomposed without replaying their original visual tokens through the model.

\subsection{Geometry-Guided Memory Retrieval}
\label{sec:geometry_retrieval}

Given the individually addressable chunk-wise memories, we retrieve historical chunks according to their geometric relevance to the current view.
For each query chunk $q$ and an eligible historical chunk $i$, we estimate their overlap using only camera poses, intrinsics, and a scene-level median-depth estimate.
For a sampled image-plane location $(u,v)$, we construct the camera-space ray
\begin{equation}
\mathbf{r}(u,v)
=
\left[
\frac{u-c_x}{f_x},
\frac{v-c_y}{f_y},
1
\right]^\top,
\end{equation}
and instantiate proxy 3D points at several depths around the scene-level median depth,
\begin{equation}
\mathcal{D}
=
\{0.5d_{\mathrm{med}},\, d_{\mathrm{med}},\, 1.5d_{\mathrm{med}}\}.
\end{equation}
These proxy points are transformed into world coordinates using the corresponding camera poses and projected into the camera views of the other chunk.

Let $V(i \rightarrow q)$ denote the fraction of proxy points constructed from chunk $i$ that project with positive depth into at least one camera view of chunk $q$.
We define the symmetric frustum-overlap score as
\begin{equation}
R(i,q)
=
\frac{1}{2}
\left[
V(i \rightarrow q)
+
V(q \rightarrow i)
\right].
\label{eq:fov_overlap_main}
\end{equation}
The score provides a lightweight proxy for geometric overlap rather than exact 3D scene intersection.
Similar geometry-based criteria have been used to retrieve relevant historical context for long-horizon video generation and world simulation~\citep{yu2025context,xiao2026worldmem}.
Importantly, our retrieval requires neither generated RGB content nor learned semantic features.
We rank the eligible historical chunks according to $R(i,q)$ and retrieve the top-$K$ most relevant ones.
Further details of the camera sampling and overlap computation are provided in Appendix~\ref{app:fov_retrieval}.

In addition to the retrieved historical chunks, we retain the first chunk as a persistent sink to preserve global context, together with recent chunks to maintain local temporal continuity. When the change in camera translation direction exceeds a threshold $\tau_{\mathrm{turn}}$, we temporarily expand the recent context to the three most recent chunks to improve robustness under abrupt camera motion.

Let $\mathcal{H}_q$ denote the remaining historical chunks after excluding the sink and recent context. The selected history is
\begin{equation}
\mathcal{I}_q
=
\mathcal{I}_{\mathrm{sink}}
\cup
\mathcal{I}_{\mathrm{recent}}
\cup
\operatorname{TopK}_{i\in\mathcal{H}_q} R(i,q).
\label{eq:memory_selection}
\end{equation}
This design combines sparse retrieval of geometrically relevant long-range memory with adaptive preservation of short-term context, reducing interference from unrelated intermediate scenes while maintaining local continuity during abrupt camera motion.

\subsection{Selective State Recomposition}
\label{sec:state_recomposition}

After retrieval, the selected chunks are restored to their original chronological order. Denote the ordered set as
\begin{equation}
\pi_q = (\pi_1,\pi_2,\ldots,\pi_M).
\end{equation}
Using the associative rule in Eq.~\ref{eq:summary_composition}, their summaries can be directly composed as
\begin{equation}
A_{\pi_q}
=
\prod_{m=1}^{M}A_{\pi_m},
\label{eq:selected_transition}
\end{equation}
and
\begin{equation}
B_{\pi_q}
=
\sum_{m=1}^{M}
B_{\pi_m}
\prod_{n=m+1}^{M}A_{\pi_n}.
\label{eq:selected_write}
\end{equation}
The resulting historical recurrent state is
\begin{equation}
S_q^{\mathrm{hist}}
=
S_0A_{\pi_q}+B_{\pi_q},
\label{eq:selected_state}
\end{equation}
where $S_0$ is the initial GDN state.

Standard GDN composes every historical chunk in temporal order, causing information written by an early scene to be repeatedly transformed by all subsequent chunks. In contrast, \methodname~shortens this propagation path by excluding geometrically irrelevant intermediate chunks. Distant but relevant memories are therefore exposed to substantially fewer unrelated transitions before being reused. Meanwhile, the original GDN dynamics within each selected chunk are fully preserved, rather than replacing them with direct averaging or summation of linear states.

The recomposition is performed directly from cached $(A_i,B_i)$ summaries and does not require re-forwarding the selected historical chunks. When all previous chunks are selected, the composition reduces to the original chronological GDN recurrence, providing a direct consistency check for the chunk-wise formulation. For the accompanying softmax-attention pathway, we apply the same selected chunk set to the corresponding historical KV cache. 

\begin{table*}[t]
\centering
\caption{
Main results on SANA-WM-Bench under different generation stages and refinement modes.
We report revisit consistency, camera controllability, and inference throughput.
Best results within each setting are shown in bold.
}
\label{tab:main_results}
\resizebox{\textwidth}{!}{
\begin{tabular}{llccccccc}
\toprule
\multirow{2}{*}{Pipeline}
& \multirow{2}{*}{Method}
& \multicolumn{3}{c}{Revisit Consistency}
& \multicolumn{3}{c}{Camera Control}
& \multicolumn{1}{c}{Efficiency} \\
\cmidrule(lr){3-5}
\cmidrule(lr){6-8}
\cmidrule(lr){9-9}
&
&
PSNR $\uparrow$
& SSIM $\uparrow$
& LPIPS $\downarrow$
& RotErr ($^\circ$) $\downarrow$
& TransErr $\downarrow$
& CamMC $\downarrow$
& FPS $\uparrow$ \\
\midrule

\multicolumn{9}{l}{\textit{Simple-Trajectory Split}} \\
\cmidrule(lr){1-9}

\multirow{2}{*}{Stage-1}
& SANA-WM
& 9.18
& 0.1729
& 0.6327
& 19.7040
& 2.2625
& 2.3974
& \textbf{22.403} \\
& \methodname
& \textbf{9.91}
& \textbf{0.1999}
& \textbf{0.6049}
& \textbf{13.5864}
& \textbf{2.0928}
& \textbf{2.1798}
& 22.053 \\

\cmidrule(lr){1-9}

\multirow{2}{*}{Stage-1 + AR Refine}
& SANA-WM
& 14.44
& 0.2770
& 0.5738
& 20.9315
& \textbf{2.1587}
& \textbf{2.3121}
& \textbf{8.280} \\
& \methodname
& \textbf{14.82}
& \textbf{0.2872}
& \textbf{0.5682}
& \textbf{20.2933}
& 2.1910
& 2.3240
& 8.211 \\

\cmidrule(lr){1-9}

\multirow{2}{*}{Stage-1 + Bi. Refine}
& SANA-WM
& 13.04
& 0.2945
& 0.5786
& 11.3848
& 1.9893
& 2.0545
& \textbf{7.701} \\
& \methodname
& \textbf{13.56}
& \textbf{0.3148}
& \textbf{0.5614}
& \textbf{8.8823}
& \textbf{1.8637}
& \textbf{1.9127}
& 7.700 \\

\midrule

\multicolumn{9}{l}{\textit{Hard-Trajectory Split}} \\
\cmidrule(lr){1-9}

\multirow{2}{*}{Stage-1}
& SANA-WM
& 9.37
& 0.1774
& 0.6044
& 20.3759
& 1.9952
& 2.1507
& \textbf{22.403} \\
& \methodname
& \textbf{10.11}
& \textbf{0.1997}
& \textbf{0.5848}
& \textbf{15.0887}
& \textbf{1.8388}
& \textbf{1.9482}
& 22.053 \\

\cmidrule(lr){1-9}

\multirow{2}{*}{Stage-1 + AR Refine}
& SANA-WM
& 14.12
& 0.2719
& 0.5621
& 25.6330
& 2.0099
& 2.2079
& \textbf{8.280} \\
& \methodname
& \textbf{14.47}
& \textbf{0.2812}
& \textbf{0.5557}
& \textbf{22.0165}
& \textbf{1.9351}
& \textbf{2.1043}
& 8.211 \\

\cmidrule(lr){1-9}

\multirow{2}{*}{Stage-1 + Bi. Refine}
& SANA-WM
& 13.05
& 0.2972
& 0.5468
& 13.1713
& 1.6649
& 1.7694
& \textbf{7.701} \\
& \methodname
& \textbf{13.51}
& \textbf{0.3142}
& \textbf{0.5384}
& \textbf{10.5268}
& \textbf{1.5449}
& \textbf{1.6230}
& 7.700 \\

\bottomrule
\end{tabular}
}
\end{table*}

\textbf{Efficient State Recomposition Kernel.}~We implement direct state recomposition on top of the chunk-wise Triton GDN kernels. During the original forward pass, the affine summaries $(A_i,B_i)$ of each chunk are computed once and cached independently of the incoming recurrent state. For each query chunk $q$, the summaries in $\mathcal{I}_q$ are gathered in chronological order and directly composed through an efficient forward scan to reconstruct $S_q^{\mathrm{hist}}$.

\textbf{Memory and Computational Overhead.}~\methodname~introduces additional storage only for a fixed-size affine summary $(A_i,B_i)$ and lightweight camera metadata for each temporal chunk. Since historical information is cached at the chunk level rather than the token level, the memory footprint scales with the number of chunks instead of the number of visual tokens. During generation, only the selected summaries are involved in state composition, while geometry-based retrieval operates directly on camera information with minimal overhead. Consequently, \methodname~provides selective access to long-range memory while largely preserving the memory and computational efficiency of recurrent linear attention.

\section{Experiments}
\label{sec:experiments}

\subsection{Experimental Setup}

\textbf{Datasets and Baselines.}~
We conduct experiments on SANA-WM-Bench~\citep{zhu2026sana} and MBench-A~\citep{zhang2026mbench}.
SANA-WM-Bench contains simple and hard splits, each with $80$ camera-controlled trajectories spanning four scene categories: indoor, outdoor city, outdoor nature, and game-style environments.
Each trajectory contains $961$ frames at $16$ fps, corresponding to approximately $60$ seconds, with a resolution of $1280\times704$.
MBench-A contains $547$ samples across four official subsets: Causal, Human, Environment, and Object.
We use the official $1.6$B SANA-WM as our baseline.
Its Stage~1 model performs $4$-step distilled autoregressive streaming generation with recurrent GDN states, followed by optional causal AR or bidirectional refinement.
\methodname~modifies only the Stage~1 memory mechanism and requires no additional training or fine-tuning, while downstream refinement remains unchanged.
Within each setting, the baseline and \methodname~use identical inputs, camera trajectories, prompts, and generation configurations.

\textbf{Implementation Details.}~
All video-generation experiments are conducted on a single NVIDIA H200 GPU with $140$ GB of memory.
We follow the official SANA-WM autoregressive streaming pipeline~\citep{zhu2026sana}, including the chunk-causal LTX-2 refiner and causal VAE.
Following the default $4$-step distilled configuration, we use the sampling schedule $[1000,960,889,727,0]$ and three latent frames per autoregressive block.
Unless otherwise specified, \methodname~retains one sink chunk, the most recent chunk, and the top-$1$ historical chunk ranked by geometric overlap.
When the within-chunk change in camera translation direction exceeds $\tau_{\mathrm{turn}}=25^\circ$, the recent context is temporarily expanded from one to three chunks.

\textbf{Evaluation Metrics.}~
We evaluate long-range scene consistency using PSNR, SSIM~\citep{wang2004image}, and LPIPS~\citep{zhang2018unreasonable} between frames corresponding to repeated or near-repeated camera poses. Camera controllability is measured by rotation error ($\mathrm{RotErr}$), relative translation error ($\mathrm{TransErr}_{\mathrm{rel}}$), and relative camera-motion consistency ($\mathrm{CamMC}_{\mathrm{rel}}$), following the SANA-WM evaluation protocol~\citep{zhu2026sana}. Camera trajectories are estimated using Pi3X, an enhanced implementation of $\pi^3$~\citep{wang2026pi}.

\subsection{Main Results}
\label{sec:main_results}

\begin{figure*}[t]
    \centering

    \begin{minipage}[t]{0.59\textwidth}
    \vspace{0pt}
    \centering
    \captionof{table}{
    Overall results on MBench-A under different inference modes.
    Best results within each setting are shown in bold.
    }
    \label{tab:mbench_results}
    \scriptsize
    \resizebox{\linewidth}{!}{%
    \begin{tabular}{llccc}
        \toprule
        Pipeline & Method
        & PSNR $\uparrow$
        & SSIM $\uparrow$
        & LPIPS $\downarrow$ \\
        \midrule

        \multirow{2}{*}{Stage-1}
        & SANA-WM
        & 10.17
        & 0.2545
        & 0.6251 \\
        & \methodname
        & \textbf{11.01}
        & \textbf{0.2848}
        & \textbf{0.5823} \\

        \midrule

        \multirow{2}{*}{Stage-1 + AR Refine}
        & SANA-WM
        & 12.36
        & 0.2996
        & 0.6554 \\
        & \methodname
        & \textbf{12.73}
        & \textbf{0.3122}
        & \textbf{0.6447} \\

        \midrule

        \multirow{2}{*}{Stage-1 + Bi. Refine}
        & SANA-WM
        & 12.94
        & 0.3310
        & 0.5597 \\
        & \methodname
        & \textbf{13.52}
        & \textbf{0.3431}
        & \textbf{0.5470} \\

        \bottomrule
    \end{tabular}}
    \end{minipage}
    \hfill
    \begin{minipage}[t]{0.40\textwidth}
        \vspace{0pt}
        \centering
        \includegraphics[width=\linewidth]{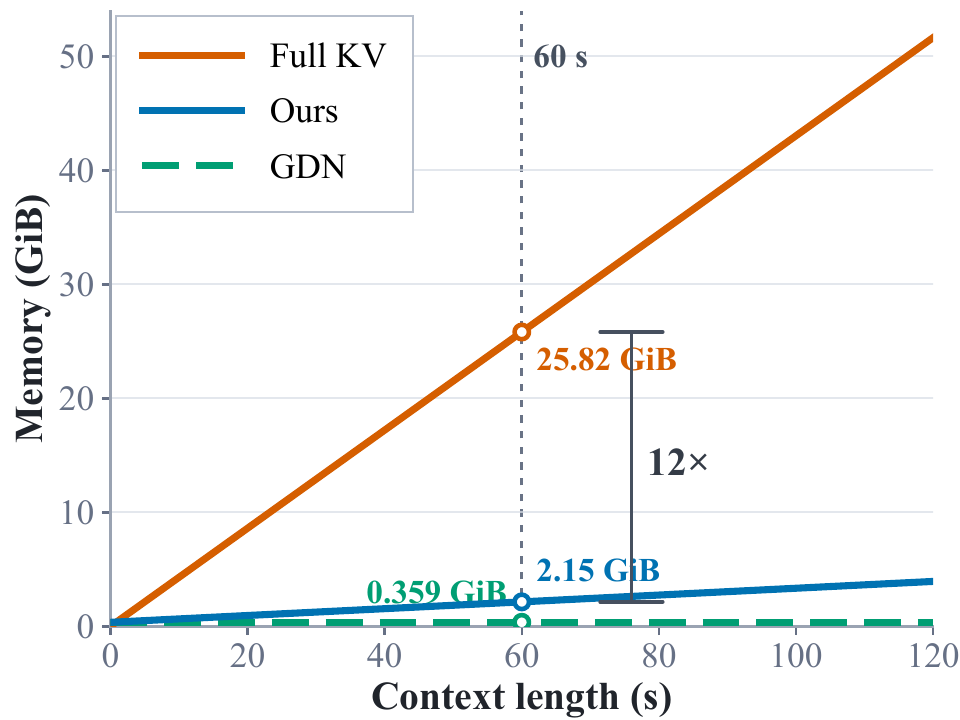}
        \vspace{-7.5mm}
        \captionof{figure}{
        Memory scaling comparison.
        }
        \label{fig:efficiency}
    \end{minipage}

\end{figure*}

\textbf{Performance on SANA-WM-Bench.}~
Table~\ref{tab:main_results} evaluates \methodname~across different stages of the SANA-WM generation pipeline.
At Stage~1, where \methodname~directly modifies the recurrent memory, it improves all revisit-consistency and camera-control metrics on both trajectory splits.
On Hard trajectories, PSNR and SSIM increase by $0.74$ dB and $0.0223$, LPIPS decreases by $0.0196$, and RotErr is reduced by $26.0\%$.
Similar improvements are observed on Simple trajectories, including a $0.73$ dB PSNR gain and a $31.0\%$ reduction in RotErr.
These camera-control gains are consistent with improved scene stability, as preserving previously observed geometry reduces structural drift and provides a more reliable spatial reference for camera-conditioned generation.
The improvements largely persist after downstream refinement.
Under causal AR refinement, all three revisit-consistency metrics improve on both splits.
On Hard trajectories, all three camera-control metrics also improve, while on Simple trajectories RotErr decreases but TransErr and CamMC increase slightly.
Full-sequence refinement preserves the benefits more consistently and improves all six metrics on both splits, including PSNR gains of $0.52$/$0.46$ dB and RotErr reductions of $22.0\%$/$20.1\%$ on the Simple/Hard splits.
Overall, selective long-range memory consistently improves revisit fidelity while generally enhancing camera controllability over long rollouts.
Detailed category-level results, together with extensive ablations on component contributions, GDN/KV alignment, retrieval budget, and historical selection strategies, are provided in the Appendix.

\textbf{Performance on MBench.}~
We further evaluate \methodname~on the full MBench-A benchmark with $547$ samples to assess its generalization beyond SANA-WM-Bench.
As shown in Table~\ref{tab:mbench_results}, \methodname~consistently improves PSNR, SSIM, and LPIPS across all three inference modes.
The gains are most pronounced at Stage~1, where PSNR and SSIM increase by $0.84$ dB and $0.0303$, respectively, while LPIPS decreases by $0.0428$.
The improvements remain consistent after downstream refinement, with PSNR gains of $0.37$ dB and $0.58$ dB under causal AR and bidirectional refinement, respectively.
These results demonstrate that the proposed memory mechanism generalizes effectively across different refinement modes and diverse scene and interaction patterns.
Detailed subset-level results are provided in the Appendix.

\textbf{Efficiency Analysis.}~
\methodname~offers a practical trade-off between full-KV attention and recurrent linear attention.
While full KV caching preserves direct access to the entire history at a memory cost that grows rapidly with context length, native GDN maintains constant memory but compresses the entire history into a single recurrent state.
\methodname~retains selectively addressable long-range memory through compact chunk-wise summaries, requiring only $2.15$ GiB of historical-state memory at a $60$-second context, $12\times$ less than full KV caching, while incurring at most a $1.6\%$ reduction in inference throughput across the evaluated settings.

\begin{figure*}[t]
    \centering
    \includegraphics[width=0.95\textwidth]
    {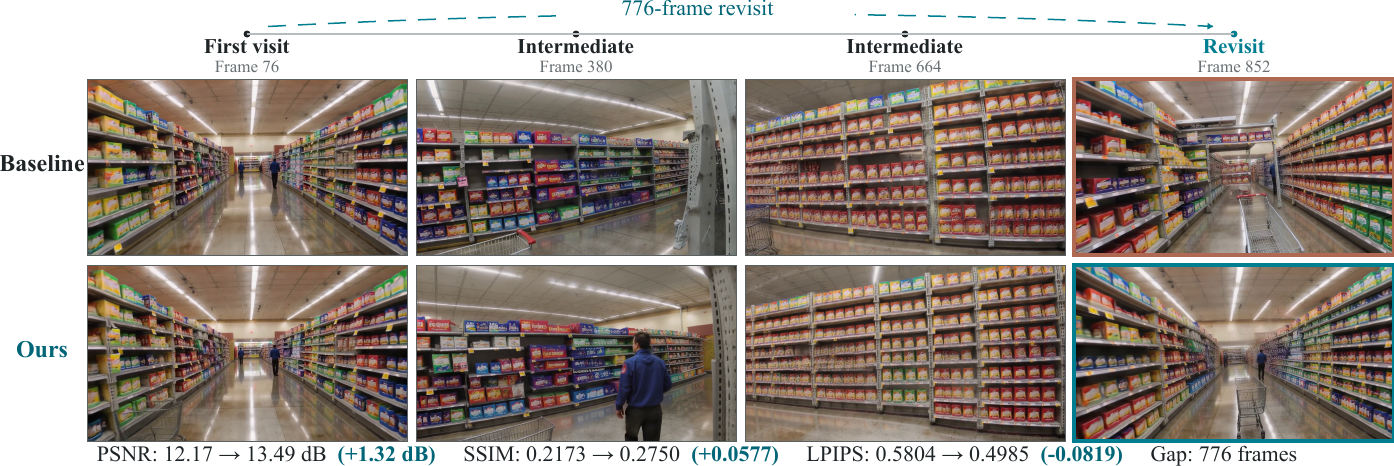}
    
    \vspace{1.5mm}
    
    \includegraphics[width=0.95\textwidth]
    {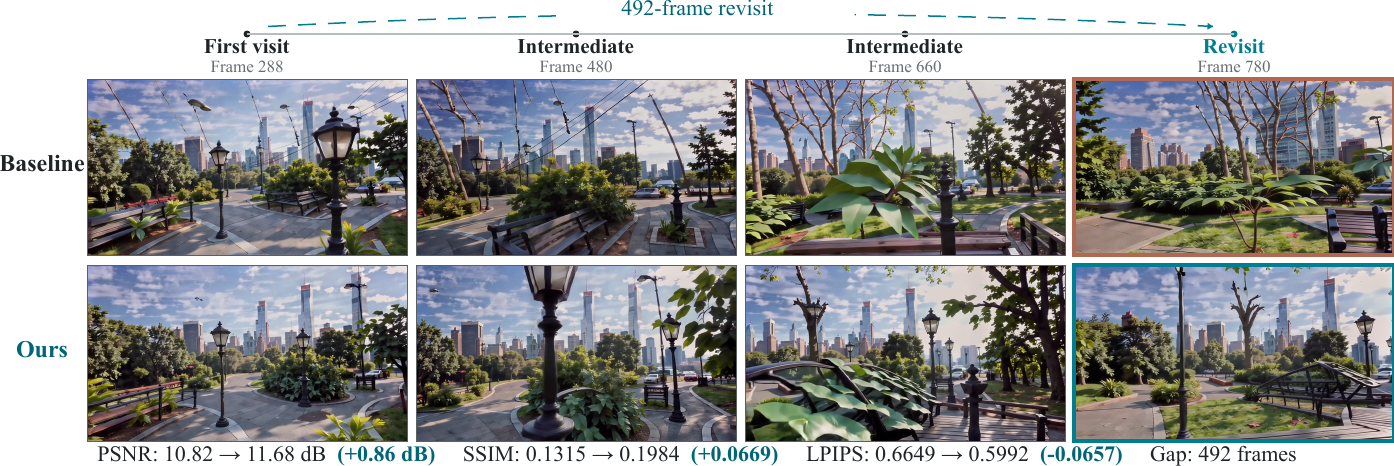}
    
    \caption{
    Qualitative comparison on long-range revisit scenarios under causal AR refinement (top) and bidirectional refinement (bottom).
    Each example shows the first visit, intermediate views, and a later revisit along the same camera trajectory.
    Revisit metrics are computed between the first-visit and revisit frames within each method.
    }
    \label{fig:qualitative}
\end{figure*}

\textbf{Qualitative Results.}~
Figure~\ref{fig:qualitative} further illustrates the long-range memory improvements of \methodname~under different refinement modes.
In the AR-refined indoor example, the baseline exhibits substantial structural drift after a long revisit: the aisle geometry changes, the shelf arrangement becomes inconsistent, and the alignment between the shelves and ceiling lights deviates noticeably from the first visit.
In contrast, \methodname~better preserves the two-sided shelf layout, central aisle structure, and overall scene perspective.
In the bidirectionally refined outdoor example, the baseline shows larger changes in the foreground composition and spatial arrangement of the walkway, vegetation, and distant buildings.
\methodname~more consistently preserves these scene elements and their relative geometry upon revisit.
Additional visualizations are provided in the Appendix.

\section{Conclusion}
\label{sec:conclusion}

We identify long-range forgetting as a key limitation of recurrent linear attention in video world models. Our chunk-level analysis of GDN shows that distant but relevant scene information is progressively attenuated by subsequent, often unrelated, state transitions. Based on this insight, we propose \methodname, which retrieves geometrically relevant chunk-wise memories and selectively recomposes them for long-range recall. Experiments on SANA-WM-Bench demonstrate consistent improvements in revisit fidelity and camera controllability, while retaining substantially lower memory cost than full KV caching and introducing only modest inference overhead.

\textbf{Limitations and Future Work.}~
\methodname~is currently training-free, which limits the achievable gains. Future work could integrate selective state retrieval into training to better preserve and exploit long-range memory. In addition, our retrieval relies solely on camera geometry and FOV overlap, without explicitly modeling occlusion, visibility, or semantic relevance. Incorporating richer or learned retrieval signals may further improve robustness in complex scenes.





\clearpage
\bibliographystyle{unsrtnat}
\bibliography{iclr2027_conference}

\clearpage
\appendix
\section{Appendix}

\subsection{Additional Implementation Details}
\label{app:implementation}

\subsubsection{Geometry-Aware Historical Retrieval}
\label{app:fov_retrieval}

For each autoregressive chunk, we estimate its geometric overlap with historical chunks from the prescribed camera trajectory. The retrieval uses only camera intrinsics, extrinsics, and a scene-level median-depth scalar from the trajectory metadata, without relying on generated RGB frames, optical flow, semantic features, or dense depth maps.

For a video with $N$ output frames and $C$ autoregressive chunks, let $a_c$ and $b_c$ denote the starting and ending frame indices assigned to chunk $c$, respectively. We compute them as
\begin{equation}
a_c = \operatorname{round}\left(\frac{cN}{C}\right), \qquad
b_c = \max\left(
a_c+1,
\operatorname{round}\left(\frac{(c+1)N}{C}\right)
\right).
\end{equation}
In our $60$-second setting, $N=961$ and $C=40$, yielding approximately $24$ camera poses per chunk.

For each camera frame, we uniformly sample a $7\times7$ image-plane grid over $5\%$--$95\%$ of the image extent. Each sampled ray is instantiated at three proxy depths,
\begin{equation}
\mathcal{D}
=
\{0.5d_{\mathrm{med}},\,d_{\mathrm{med}},\,1.5d_{\mathrm{med}}\},
\end{equation}
where $d_{\mathrm{med}}$ denotes the scene-level median-depth proxy. For a pixel coordinate $(u,v)$, the corresponding camera-space ray is
\begin{equation}
\mathbf{r}(u,v)
=
\left[
\frac{u-c_x}{f_x},
\frac{v-c_y}{f_y},
1
\right]^\top.
\end{equation}
Each proxy point at depth $d\in\mathcal{D}$ is then transformed into world coordinates using the camera-to-world transformation.

Let $\mathcal{P}_i$ denote the set of proxy points constructed from chunk $i$. We define the directed visibility score $V(i\rightarrow j)$ as the fraction of points in $\mathcal{P}_i$ that project with positive depth into the image plane of at least one camera frame in chunk $j$. The symmetric overlap score is then
\begin{equation}
R(i,j)
=
\frac{1}{2}
\left[
V(i\rightarrow j)+V(j\rightarrow i)
\right].
\label{eq:fov_overlap}
\end{equation}
This score provides a lightweight proxy for frustum overlap rather than exact $3$D scene intersection, as it does not explicitly model occlusion or scene geometry.

For each query chunk $c$, we retain the sink and recent chunks and retrieve the top-$K$ remaining historical chunks according to $R(i,c)$. The selected chunks are reordered chronologically before GDN state recomposition.

\subsubsection{Turn-Aware Recent Context}
\label{app:turn_aware}

To preserve local continuity under abrupt trajectory changes, we adapt the recent-history budget based on the motion within the current query chunk. For a regular $24$-frame camera chunk with translation vectors $\{\mathbf{p}_0,\ldots,\mathbf{p}_{23}\}$, we define
\begin{equation}
\mathbf{d}_{\mathrm{early}}
=
\mathbf{p}_7-\mathbf{p}_0,
\qquad
\mathbf{d}_{\mathrm{late}}
=
\mathbf{p}_{23}-\mathbf{p}_{16},
\end{equation}
and compute their angular difference as
\begin{equation}
\theta_c
=
\arccos
\frac{
\mathbf{d}_{\mathrm{early}}^\top
\mathbf{d}_{\mathrm{late}}
}{
\|\mathbf{d}_{\mathrm{early}}\|_2
\|\mathbf{d}_{\mathrm{late}}\|_2
}.
\label{eq:turn_angle}
\end{equation}
The recent-history budget is then set to
\begin{equation}
r_c =
\begin{cases}
3, & \theta_c \ge \tau_{\mathrm{turn}},\\
1, & \text{otherwise},
\end{cases}
\label{eq:turn_budget}
\end{equation}
where $\tau_{\mathrm{turn}}=25^\circ$. The expanded recent context is applied only to the current chunk; subsequent chunks revert to the default budget unless the criterion is triggered again.

\subsection{Implementation of State Recomposition}
\label{app:gdn_recomposition}

The affine state composition described in the main text is implemented directly on top of the original GDN kernels. After each chunk is completed, we cache its frame-wise Phase-A sufficient statistics, which characterize its affine effect on an arbitrary incoming recurrent state. For each query chunk, the statistics of the selected historical chunks are gathered in chronological order and passed to the original Phase-B forward scan to reconstruct the query-specific GDN state. This avoids re-evaluating historical chunks through the Transformer.

Selective recomposition is applied only to the main GDN matrix and normalization states. The camera-GDN recurrent state, GDN short-convolution state, and FFN temporal-convolution state retain their original chronological updates. We additionally batch compatible GDN blocks during recomposition to reduce kernel-launch and data-movement overhead without changing the underlying recurrence.

\subsubsection{Overall Inference Procedure}
\label{app:overall_algorithm}

Algorithm~\ref{alg:overall} summarizes the complete inference procedure of \methodname.
The pairwise geometric overlap scores are first computed from the prescribed camera trajectory using Eq.~\ref{eq:fov_overlap}.
At each autoregressive step, we determine the recent-history budget according to the within-chunk trajectory change, select the historical chunks using Eq.~\ref{eq:memory_selection}, and recompose the corresponding main-GDN state.
The current chunk is then generated using the resulting query-specific memory.
After generation, an additional $t=0$ forward pass caches the affine summary of the completed chunk for subsequent retrieval.
Throughout this process, camera-control and local convolutional states retain their original chronological updates.

\begin{algorithm}[t]
\caption{Overall Inference Procedure of \methodname}
\label{alg:overall}
\begin{algorithmic}[1]
\Require Prescribed camera trajectory, retrieval budget $K$, and turn threshold $\tau_{\mathrm{turn}}$
\State Compute pairwise geometric overlap scores $R(i,j)$ using Eq.~\ref{eq:fov_overlap}
\State Initialize the chunk-wise GDN summary bank and streaming states

\For{query chunk $q=0,\ldots,C-1$}
    \State Compute the recent-history budget $r_q$ using Eqs.~\ref{eq:turn_angle}--\ref{eq:turn_budget}

    \If{insufficient history for selective retrieval}
        \State $\mathcal{I}_q \gets \{0,\ldots,q-1\}$
    \Else
        \State Select historical chunks $\mathcal{I}_q$ according to Eq.~\ref{eq:memory_selection}
        \State Sort $\mathcal{I}_q$ in chronological order
        \State Gather the cached affine summaries of chunks in $\mathcal{I}_q$
        \State Recompose the query-specific historical state $S_q^{\mathrm{hist}}$
        \State Replace the main-GDN recurrent state with $S_q^{\mathrm{hist}}$
    \EndIf

    \State Generate chunk $q$ using the resulting streaming cache
    \State Perform the $t=0$ cache-writing forward pass
    \State Cache the affine summary of chunk $q$
    \State Update camera-GDN and local convolutional states chronologically
\EndFor
\end{algorithmic}
\end{algorithm}

The retrieval procedure accesses only historical chunks, and the selected chunks are always recomposed in their original temporal order.
Importantly, \methodname~recomposes only the main-GDN recurrent memory; the camera-GDN recurrent state and local convolutional states follow the original streaming recurrence.
Historical chunks are therefore never re-evaluated through the Transformer.

\subsection{Component Ablation}

Table~\ref{tab:component_ablation} evaluates the major components of \methodname~with $K=1$ on the Hard-Trajectory split.
Here, ``w/o Select $A$'' retains the transition terms $A_i$ of all historical chunks while selectively composing their write terms, whereas ``w/o Select $B$'' retains all historical write terms $B_i$ while selectively composing the transitions.
Removing the sink degrades both revisit consistency and camera controllability, highlighting the importance of persistent global context.
Retaining the transition terms of unselected chunks substantially reduces PSNR and SSIM and increases camera errors, supporting our observation that irrelevant intermediate transitions contribute to long-range forgetting.
Similarly, retaining all historical write terms consistently degrades both revisit fidelity and camera controllability, indicating that selective composition is beneficial for both transition and write components of the recurrent memory.
Removing the turn-aware recent context also reduces revisit consistency and worsens camera control, confirming its role in preserving local continuity under abrupt camera motion.
Overall, the full \methodname~achieves the best performance across all six metrics, demonstrating the complementary contributions of selective state composition, persistent sink memory, and adaptive recent context.

\begin{table}[h]
\centering
\caption{
Component ablation on the Hard-Trajectory split using Stage~1 outputs.
Best results are shown in bold.
}
\label{tab:component_ablation}
\resizebox{0.95\linewidth}{!}{
\begin{tabular}{lcccccc}
\toprule
\multirow{2}{*}{Variant}
& \multicolumn{3}{c}{Revisit Consistency}
& \multicolumn{3}{c}{Camera Control} \\
\cmidrule(lr){2-4}
\cmidrule(lr){5-7}
&
PSNR $\uparrow$
& SSIM $\uparrow$
& LPIPS $\downarrow$
& RotErr ($^\circ$) $\downarrow$
& TransErr $\downarrow$
& CamMC $\downarrow$ \\
\midrule

\methodname
& \textbf{10.11}
& \textbf{0.1997}
& \textbf{0.5848}
& \textbf{15.0887}
& \textbf{1.8388}
& \textbf{1.9482} \\

\midrule

w/o Sink
& 9.71
& 0.1867
& 0.6019
& 15.9351
& 1.9326
& 2.0531 \\

w/o Select $A$
& 9.61
& 0.1887
& 0.5901
& 15.2426
& 1.8552
& 2.0562 \\

w/o Select $B$
& 9.64
& 0.1892
& 0.5921
& 15.2326
& 1.9830
& 2.0564 \\

w/o Turn-aware
& 9.96
& 0.1917
& 0.5876
& 15.1784
& 2.1096
& 1.9899 \\

\bottomrule
\end{tabular}
}
\end{table}

\subsection{GDN-State Recomposition and KV-Cache Alignment}
\label{sec:gdn_kv_ablation}

We further isolate the roles of GDN-state recomposition and softmax KV-cache alignment under the Top-$1$ retrieval setting.
\textit{GDN-only} applies selective recomposition to the main GDN memory while retaining the native softmax KV history.
\textit{KV-only} applies the selected history only to the softmax KV cache while keeping the native GDN states.
The full \methodname~applies the same retrieved history to both memory pathways.
Tables~\ref{tab:gdn_kv_ablation} and~\ref{tab:gdn_kv_ablation_mbench} report the results on SANA-WM-Bench and MBench-A, respectively.

\begin{table*}[t]
\centering
\small
\setlength{\tabcolsep}{4.0pt}
\caption{
Ablation of GDN-state recomposition and KV-cache alignment on SANA-WM-Bench under Top-$1$ retrieval.
Best results within each split and inference setting are shown in bold.
}
\label{tab:gdn_kv_ablation}
\resizebox{\textwidth}{!}{%
\begin{tabular}{llcccccc}
\toprule
Pipeline & Method
& PSNR $\uparrow$
& SSIM $\uparrow$
& LPIPS $\downarrow$
& RotErr ($^\circ$) $\downarrow$
& TransErr $\downarrow$
& CamMC $\downarrow$ \\
\midrule

\multicolumn{8}{c}{\textit{Simple-Trajectory Split}} \\
\midrule

Stage~1
& Baseline
& 9.18 & 0.1729 & 0.6327 & 19.7040 & 2.2625 & 2.3974 \\
& GDN-only
& 9.45 & 0.1812 & 0.6219 & 17.9934 & 2.2832 & 2.4083 \\
& KV-only
& 9.50 & 0.1866 & 0.6149 & 14.4283 & \textbf{1.9569} & \textbf{2.0556} \\
& \methodname
& \textbf{9.91} & \textbf{0.1999} & \textbf{0.6049} & \textbf{13.5864} & 2.0928 & 2.1798 \\
\midrule

Stage~1 + AR Refine
& Baseline
& 14.44 & 0.2770 & 0.5738 & 20.9315 & \textbf{2.1587} & \textbf{2.3121} \\
& GDN-only
& 14.65 & 0.2784 & 0.5687 & 21.2344 & 2.3596 & 2.5147 \\
& KV-only
& 13.88 & 0.2703 & 0.5812 & 20.9084 & 2.2087 & 2.3662 \\
& \methodname
& \textbf{14.82} & \textbf{0.2872} & \textbf{0.5682} & \textbf{20.2933} & 2.1910 & 2.3240 \\
\midrule

Stage~1 + Bi. Refine
& Baseline
& 13.04 & 0.2945 & 0.5786 & 11.3848 & 1.9893 & 2.0545 \\
& GDN-only
& 13.26 & 0.2998 & 0.5746 & 10.7564 & \textbf{1.8627} & 1.9310 \\
& KV-only
& 13.17 & 0.2999 & 0.5762 & 9.6939 & 1.8866 & 1.9417 \\
& \methodname
& \textbf{13.56} & \textbf{0.3148} & \textbf{0.5614} & \textbf{8.8823} & 1.8637 & \textbf{1.9127} \\

\midrule
\multicolumn{8}{c}{\textit{Hard-Trajectory Split}} \\
\midrule

Stage~1
& Baseline
& 9.37 & 0.1774 & 0.6044 & 20.3759 & 1.9952 & 2.1507 \\
& GDN-only
& 9.78 & 0.1890 & 0.5929 & 20.0838 & 1.9502 & 2.1075 \\
& KV-only
& 9.52 & 0.1805 & 0.6000 & \textbf{13.3992} & \textbf{1.7635} & \textbf{1.8616} \\
& \methodname
& \textbf{10.11} & \textbf{0.1997} & \textbf{0.5848} & 15.0887 & 1.8388 & 1.9482 \\
\midrule

Stage~1 + AR Refine
& Baseline
& 14.12 & 0.2719 & 0.5621 & 25.6330 & 2.0099 & 2.2079 \\
& GDN-only
& 14.35 & 0.2797 & 0.5576 & 23.1860 & 1.9519 & 2.1296 \\
& KV-only
& 13.54 & 0.2599 & 0.5721 & 24.1145 & 2.0425 & 2.2302 \\
& \methodname
& \textbf{14.47} & \textbf{0.2812} & \textbf{0.5557} & \textbf{22.0165} & \textbf{1.9351} & \textbf{2.1043} \\
\midrule

Stage~1 + Bi. Refine
& Baseline
& 13.05 & 0.2972 & 0.5468 & 13.1713 & 1.6649 & 1.7694 \\
& GDN-only
& 13.49 & \textbf{0.3304} & 0.5601 & 33.0406 & 2.1843 & 2.4584 \\
& KV-only
& 13.26 & 0.3047 & 0.5469 & 11.4688 & 1.5860 & 1.6696 \\
& \methodname
& \textbf{13.51} & 0.3142 & \textbf{0.5384} & \textbf{10.5268} & \textbf{1.5449} & \textbf{1.6230} \\

\bottomrule
\end{tabular}%
}
\end{table*}

\begin{table}[t]
\centering
\small
\setlength{\tabcolsep}{4.5pt}
\caption{
Ablation of GDN-state recomposition and KV-cache alignment on MBench-A under Top-$1$ retrieval.
Best results within each inference setting are shown in bold.
}
\label{tab:gdn_kv_ablation_mbench}
\begin{tabular}{llccc}
\toprule
Pipeline & Method
& PSNR $\uparrow$
& SSIM $\uparrow$
& LPIPS $\downarrow$ \\
\midrule

\multirow[c]{4}{*}{Stage~1}
& Baseline & 10.17 & 0.2545 & 0.6251 \\
& GDN-only & 10.61 & 0.2626 & 0.6052 \\
& KV-only & 10.43 & 0.2606 & 0.6091 \\
& \methodname & \textbf{11.01} & \textbf{0.2848} & \textbf{0.5823} \\
\midrule

\multirow[c]{4}{*}{Causal AR Refine}
& Baseline & 12.36 & 0.2996 & 0.6554 \\
& GDN-only & 12.57 & 0.3064 & 0.6492 \\
& KV-only & 12.35 & 0.3016 & 0.6547 \\
& \methodname & \textbf{12.73} & \textbf{0.3122} & \textbf{0.6447} \\
\midrule

\multirow[c]{4}{*}{Bidirectional Refine}
& Baseline & 12.94 & 0.3310 & 0.5597 \\
& GDN-only & 13.28 & 0.3389 & 0.5477 \\
& KV-only & 13.04 & 0.3325 & 0.5563 \\
& \methodname & \textbf{13.52} & \textbf{0.3431} & \textbf{0.5470} \\
\bottomrule
\end{tabular}
\end{table}

The two memory pathways play complementary but asymmetric roles.
At Stage~1, GDN-only already yields substantial gains in revisit consistency on both SANA-WM-Bench splits, whereas KV-only provides comparatively smaller improvements.
Their combination further improves long-range recall, with the full \methodname~achieving the best PSNR, SSIM, and LPIPS on both splits.
KV alignment additionally improves camera stability, although its effect varies across trajectory types.
Under causal AR refinement, the full method achieves the strongest revisit performance on both splits and improves all six metrics on Hard trajectories.
These results indicate that GDN-state recomposition is the primary source of long-range memory improvement, while KV-cache alignment provides complementary stabilization of the attention-based context.

The bidirectional setting further illustrates the interaction between the two pathways.
On Hard trajectories, GDN-only substantially improves SSIM but increases RotErr from $13.17^\circ$ to $33.04^\circ$.
This suggests that modifying only the recurrent history can create a mismatch with the accompanying softmax context.
Applying the same historical selection to the KV cache restores consistency between the two memory pathways, and the full method achieves the best PSNR, LPIPS, and all three camera-control metrics.
Thus, KV alignment is particularly important for maintaining stable generation when the recurrent history is selectively recomposed.

MBench-A exhibits an even clearer pattern.
Across Stage~1, causal AR refinement, and bidirectional refinement, GDN-only consistently provides larger gains than KV-only, while the full \methodname~achieves the best PSNR, SSIM, and LPIPS in every setting.
This confirms that directly modifying the recurrent GDN memory is the main driver of improved long-range recall, whereas KV-cache alignment alone provides limited gains.
Combining the two yields the strongest and most consistent overall performance.

\subsection{Number of Retrieved Historical Chunks}

We study the effect of the retrieval budget $K$ on the Hard-Trajectory split using Stage~1 outputs.
All variants use the same sink memory and turn-aware recent-context strategy, differing only in the number of historical chunks retrieved by geometric overlap.
As shown in Table~\ref{tab:ablation_topk}, performance remains broadly comparable across different retrieval budgets, and increasing $K$ does not consistently improve either revisit fidelity or camera controllability.
Notably, $K=1$ achieves the best PSNR, SSIM, and LPIPS while using the smallest retrieval budget.
Larger values provide modest gains in some camera-control metrics, with $K=5$ achieving the lowest RotErr and $K=4$ the lowest TransErr and CamMC.
These results indicate that a single highly relevant historical chunk is often sufficient for effective long-range recall.
We therefore adopt $K=1$ as the default setting for its favorable trade-off between retrieval quality and efficiency.

\begin{table}[h]
\centering
\caption{
Ablation on the retrieval budget $K$ on the Hard-Trajectory split using Stage~1 outputs.
Best results are shown in bold.
}
\label{tab:ablation_topk}
\resizebox{0.8\linewidth}{!}{
\begin{tabular}{lcccccc}
\toprule
\multirow{2}{*}{$K$}
& \multicolumn{3}{c}{Revisit Consistency}
& \multicolumn{3}{c}{Camera Control} \\
\cmidrule(lr){2-4}
\cmidrule(lr){5-7}
&
PSNR $\uparrow$
& SSIM $\uparrow$
& LPIPS $\downarrow$
& RotErr ($^\circ$) $\downarrow$
& TransErr $\downarrow$
& CamMC $\downarrow$ \\
\midrule

$1$
& \textbf{10.11}
& \textbf{0.1997}
& \textbf{0.5848}
& 15.0887
& 1.8388
& 1.9482 \\

$3$
& 10.06
& 0.1973
& 0.5865
& 14.1537
& 1.7658
& 1.8661 \\

$4$
& 10.00
& 0.1936
& 0.5885
& 12.2795
& \textbf{1.6985}
& \textbf{1.7832} \\

$5$
& 9.89
& 0.1886
& 0.5915
& \textbf{11.5435}
& 1.7200
& 1.7970 \\

\bottomrule
\end{tabular}
}
\end{table}

\subsection{Effectiveness of Geometry-Guided Historical Selection}
\label{sec:geometry_selection_ablation}

We evaluate whether geometry provides an effective signal for selecting historical states by comparing \methodname~with three geometry-agnostic selection strategies on Hard80 under Stage~1 inference.
To more clearly expose the differences among selection strategies, we conduct this comparison with a larger retrieval budget of $K=4$, allowing each method to operate over a broader historical context.
\textit{Random} deterministically samples up to six states from the available history, \textit{Recent} retains the six most recent states, and \textit{Uniform} selects six states approximately uniformly over the entire history.
In contrast, \methodname~retains a sink state and recent context while selecting distant historical chunks according to camera-frustum overlap.
All variants use the same model, generation setting, seed, and evaluation protocol, covering $80$ scenes and $400$ revisit pairs.
The three heuristic baselines use a fixed budget of at most six historical states, whereas \methodname~normally uses six states but temporarily expands the recent context when the turn-aware criterion is triggered.
We therefore regard this experiment as a comparison of historical selection strategies rather than a strictly equal-compute ablation.

\begin{table}[t]
\centering
\caption{
Comparison of historical state selection strategies on Hard80 under Stage~1 inference with $K=4$.
Higher PSNR and SSIM are better, while lower LPIPS and camera errors are better.
}
\label{tab:geometry_selection_ablation}
\resizebox{\linewidth}{!}{
\begin{tabular}{lcccccc}
\toprule
Selection Policy
& PSNR $\uparrow$
& SSIM $\uparrow$
& LPIPS $\downarrow$
& RotErr ($^\circ$) $\downarrow$
& TransErr $\downarrow$
& CamMC $\downarrow$ \\
\midrule
Random
& 9.70
& 0.1864
& 0.6049
& 14.1883
& 1.7822
& 1.8924 \\

Recent
& 9.29
& 0.1696
& 0.6179
& 27.6876
& 2.0142
& 2.2408 \\

Uniform
& 9.91
& 0.1910
& 0.5957
& 12.3324
& \textbf{1.6881}
& 1.7838 \\

\methodname
& \textbf{10.00}
& \textbf{0.1936}
& \textbf{0.5885}
& \textbf{12.2795}
& 1.6985
& \textbf{1.7832} \\
\bottomrule
\end{tabular}
}
\end{table}

As shown in Table~\ref{tab:geometry_selection_ablation}, uniformly covering the history already provides a substantially stronger baseline than random or purely recent selection, highlighting the importance of maintaining access to distant context.
Uniform selection remains competitive, likely because the one-minute evaluation horizon allows uniformly sampled states to cover much of the observed scene history.
Nevertheless, geometry-guided selection further improves all three revisit-consistency metrics, increasing PSNR from $9.91$ to $10.00$ and SSIM from $0.1910$ to $0.1936$, while reducing LPIPS from $0.5957$ to $0.5885$.
It also achieves slightly lower RotErr and CamMC, while Uniform obtains a marginally lower TransErr.
These results indicate that explicitly selecting view-relevant historical states provides additional benefit beyond broad temporal coverage, supporting camera geometry as an effective criterion for long-range memory retrieval.

\subsection{Effect of Revisit Interval}
\label{app:revisit_length}

\begin{figure*}[t]
    \centering

    \includegraphics[width=0.32\textwidth]
    {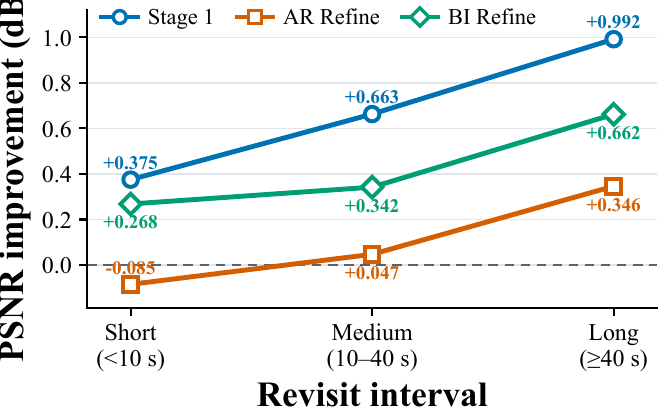}
    \hfill
    \includegraphics[width=0.32\textwidth]
    {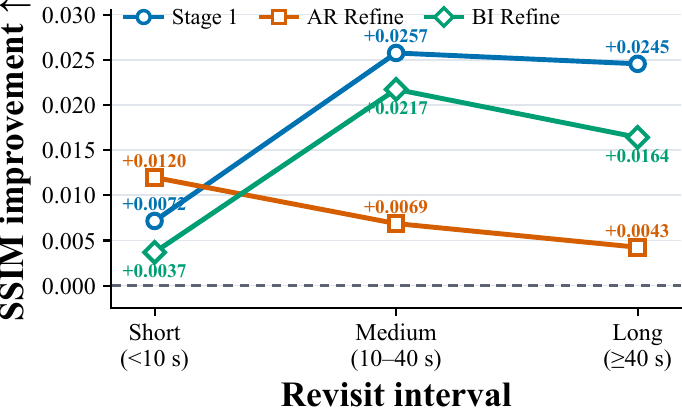}
    \hfill
    \includegraphics[width=0.32\textwidth]
    {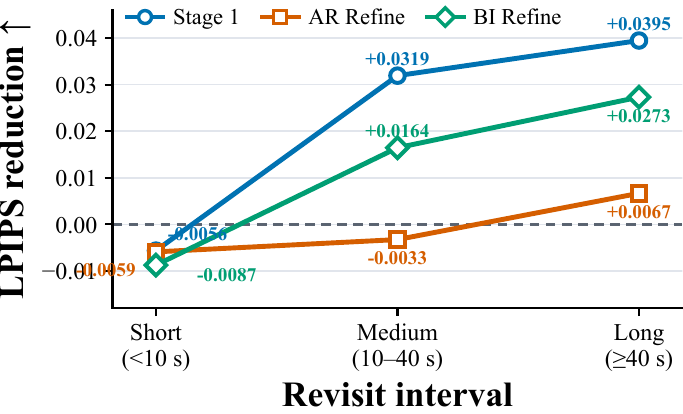}

    \caption{
    Performance improvement over SANA-WM across different revisit intervals on the Hard-Trajectory split.
    We report PSNR improvement (left), SSIM improvement (middle), and LPIPS reduction (right) for Stage~1, AR refinement, and bidirectional refinement.
    }
    \label{fig:revisit_length}
\end{figure*}

Figure~\ref{fig:revisit_length} analyzes the effect of revisit interval on the improvement of \methodname.
We group revisit pairs into short ($<10$ s), medium ($10$--$40$ s), and long ($\geq40$ s) intervals.
Across Stage~1, AR refinement, and bidirectional refinement, the gains are generally larger for medium and long revisits.
In particular, PSNR improvement increases consistently with the revisit interval across all three generation modes.
LPIPS exhibits a similar trend, shifting from minor degradation at short intervals to clear improvement for long revisits.
SSIM also shows substantially larger gains at medium or long intervals for Stage~1 and bidirectional refinement, although the trend under AR refinement is not monotonic.
Overall, these results indicate that the benefit of selective long-range memory becomes more pronounced as the temporal distance between repeated observations increases, consistent with its effectiveness in mitigating long-horizon forgetting.

\subsection{Detailed Benchmark Results}
\label{app:detailed_results}

We provide detailed category- and subset-level results to complement the aggregate results in the main paper.
We first report the four scene categories of SANA-WM-Bench, followed by the four official subsets of MBench-A.

\subsubsection{SANA-WM-Bench Category Results}
\label{app:sanawm_subsets}

SANA-WM-Bench contains four scene categories---Game Style, Indoor, Outdoor City, and Outdoor Nature---with $20$ scenes per category in each trajectory split.
Tables~\ref{tab:sanawm_stage1_categories}--\ref{tab:sanawm_bi_categories} report category-level results under Stage~1, causal AR refinement, and bidirectional refinement.
The gains are broadly distributed across scene categories, with Stage~1 showing the most consistent improvements and downstream refinement introducing several category-dependent trade-offs.

\begin{table*}[t]
\centering
\caption{
Category-level results on SANA-WM-Bench using Stage~1 outputs.
Best results within each category are shown in bold.
}
\label{tab:sanawm_stage1_categories}
\resizebox{\textwidth}{!}{
\begin{tabular}{llcccccc}
\toprule
Category & Method
& PSNR $\uparrow$
& SSIM $\uparrow$
& LPIPS $\downarrow$
& RotErr ($^\circ$) $\downarrow$
& TransErr $\downarrow$
& CamMC $\downarrow$ \\
\midrule

\multicolumn{8}{l}{\textit{Simple-Trajectory Split}} \\
\cmidrule(lr){1-8}

\multirow{2}{*}{Game Style}
& SANA-WM
& 9.85 & 0.2070 & 0.6197 & 16.5064 & \textbf{1.9242} & \textbf{2.0482} \\
& \methodname
& \textbf{11.06} & \textbf{0.2639} & \textbf{0.5866} & \textbf{12.8640} & 2.0047 & 2.0939 \\

\multirow{2}{*}{Indoor}
& SANA-WM
& 9.38 & 0.2105 & 0.6576 & 38.6940 & 1.6993 & 2.0227 \\
& \methodname
& \textbf{10.05} & \textbf{0.2338} & \textbf{0.6306} & \textbf{29.0323} & \textbf{1.6196} & \textbf{1.8481} \\

\multirow{2}{*}{Outdoor City}
& SANA-WM
& 8.96 & 0.1323 & 0.6390 & 12.2424 & 2.0782 & 2.1372 \\
& \methodname
& \textbf{9.25} & \textbf{0.1561} & \textbf{0.6240} & \textbf{5.9851} & \textbf{1.5976} & \textbf{1.6170} \\

\multirow{2}{*}{Outdoor Nature}
& SANA-WM
& 8.52 & 0.1419 & 0.6145 & 11.3733 & 3.3483 & 3.3814 \\
& \methodname
& \textbf{9.29} & \textbf{0.1459} & \textbf{0.5785} & \textbf{6.4643} & \textbf{3.1493} & \textbf{3.1603} \\

\midrule
\multicolumn{8}{l}{\textit{Hard-Trajectory Split}} \\
\cmidrule(lr){1-8}

\multirow{2}{*}{Game Style}
& SANA-WM
& 9.84 & 0.2023 & 0.5823 & \textbf{14.2161} & 1.8745 & 1.9719 \\
& \methodname
& \textbf{11.41} & \textbf{0.2550} & \textbf{0.5505} & 14.3713 & \textbf{1.8250} & \textbf{1.9349} \\

\multirow{2}{*}{Indoor}
& SANA-WM
& 9.22 & 0.2085 & 0.6390 & 36.5361 & 1.2446 & 1.6297 \\
& \methodname
& \textbf{9.63} & \textbf{0.2305} & \textbf{0.6330} & \textbf{26.6061} & \textbf{1.0675} & \textbf{1.3277} \\

\multirow{2}{*}{Outdoor City}
& SANA-WM
& 9.17 & 0.1463 & 0.6206 & 16.0176 & 1.8409 & 1.9254 \\
& \methodname
& \textbf{9.56} & \textbf{0.1611} & \textbf{0.6122} & \textbf{9.9119} & \textbf{1.6669} & \textbf{1.7070} \\

\multirow{2}{*}{Outdoor Nature}
& SANA-WM
& 9.23 & \textbf{0.1526} & 0.5756 & 14.7337 & 3.0208 & 3.0760 \\
& \methodname
& \textbf{9.84} & 0.1521 & \textbf{0.5437} & \textbf{9.4654} & \textbf{2.7958} & \textbf{2.8233} \\

\bottomrule
\end{tabular}
}
\end{table*}

\begin{table*}[t]
\centering
\caption{
Category-level results on SANA-WM-Bench after causal AR refinement.
Best results within each category are shown in bold.
}
\label{tab:sanawm_ar_categories}
\resizebox{\textwidth}{!}{
\begin{tabular}{llcccccc}
\toprule
Category & Method
& PSNR $\uparrow$
& SSIM $\uparrow$
& LPIPS $\downarrow$
& RotErr ($^\circ$) $\downarrow$
& TransErr $\downarrow$
& CamMC $\downarrow$ \\
\midrule

\multicolumn{8}{l}{\textit{Simple-Trajectory Split}} \\
\cmidrule(lr){1-8}

\multirow{2}{*}{Game Style}
& SANA-WM
& 15.19 & 0.3225 & 0.5582 & 19.8894 & 2.1729 & 2.3106 \\
& \methodname
& \textbf{15.79} & \textbf{0.3505} & \textbf{0.5492} & \textbf{16.5280} & \textbf{2.0506} & \textbf{2.1496} \\

\multirow{2}{*}{Indoor}
& SANA-WM
& 13.64 & 0.3127 & 0.6013 & 35.3855 & \textbf{1.5308} & \textbf{1.8652} \\
& \methodname
& \textbf{14.08} & \textbf{0.3172} & \textbf{0.5956} & \textbf{34.3931} & 1.6263 & 1.9040 \\

\multirow{2}{*}{Outdoor City}
& SANA-WM
& 14.21 & 0.2353 & 0.5803 & 15.1111 & 1.9558 & 2.0376 \\
& \methodname
& \textbf{14.26} & \textbf{0.2510} & \textbf{0.5777} & \textbf{14.4444} & \textbf{1.9104} & \textbf{1.9801} \\

\multirow{2}{*}{Outdoor Nature}
& SANA-WM
& 14.70 & \textbf{0.2375} & 0.5555 & \textbf{13.3400} & \textbf{2.9753} & \textbf{3.0350} \\
& \methodname
& \textbf{15.13} & 0.2302 & \textbf{0.5503} & 15.8077 & 3.1768 & 3.2625 \\

\midrule
\multicolumn{8}{l}{\textit{Hard-Trajectory Split}} \\
\cmidrule(lr){1-8}

\multirow{2}{*}{Game Style}
& SANA-WM
& 14.95 & 0.3220 & 0.5269 & 23.5136 & \textbf{1.8901} & 2.0707 \\
& \methodname
& \textbf{15.73} & \textbf{0.3398} & \textbf{0.5140} & \textbf{18.0932} & 1.8923 & \textbf{2.0223} \\

\multirow{2}{*}{Indoor}
& SANA-WM
& \textbf{13.81} & 0.3003 & 0.5923 & 35.8487 & 1.2361 & 1.6201 \\
& \methodname
& 13.71 & \textbf{0.3139} & \textbf{0.5920} & \textbf{30.6444} & \textbf{1.1457} & \textbf{1.4865} \\

\multirow{2}{*}{Outdoor City}
& SANA-WM
& 13.56 & 0.2507 & \textbf{0.5812} & 20.0231 & 1.8331 & 1.9407 \\
& \methodname
& \textbf{13.74} & \textbf{0.2570} & 0.5877 & \textbf{19.7262} & \textbf{1.8103} & \textbf{1.9228} \\

\multirow{2}{*}{Outdoor Nature}
& SANA-WM
& 14.18 & \textbf{0.2147} & 0.5478 & 23.1465 & 3.0804 & 3.2001 \\
& \methodname
& \textbf{14.71} & 0.2143 & \textbf{0.5290} & \textbf{19.6023} & \textbf{2.8921} & \textbf{2.9858} \\

\bottomrule
\end{tabular}
}
\end{table*}

\begin{table*}[t]
\centering
\caption{
Category-level results on SANA-WM-Bench after bidirectional refinement.
Best results within each category are shown in bold.
}
\label{tab:sanawm_bi_categories}
\resizebox{\textwidth}{!}{
\begin{tabular}{llcccccc}
\toprule
Category & Method
& PSNR $\uparrow$
& SSIM $\uparrow$
& LPIPS $\downarrow$
& RotErr ($^\circ$) $\downarrow$
& TransErr $\downarrow$
& CamMC $\downarrow$ \\
\midrule

\multicolumn{8}{l}{\textit{Simple-Trajectory Split}} \\
\cmidrule(lr){1-8}

\multirow{2}{*}{Game Style}
& SANA-WM
& 13.74
& 0.3351
& 0.5485
& 9.7648
& \textbf{1.7510}
& \textbf{1.8012} \\
& \methodname
& \textbf{14.20}
& \textbf{0.3553}
& \textbf{0.5377}
& \textbf{6.8672}
& 1.8842
& 1.9156 \\

\multirow{2}{*}{Indoor}
& SANA-WM
& 12.67
& 0.3174
& 0.6045
& 21.6798
& 1.4978
& 1.6675 \\
& \methodname
& \textbf{13.03}
& \textbf{0.3401}
& \textbf{0.5980}
& \textbf{16.1626}
& \textbf{1.0450}
& \textbf{1.1800} \\

\multirow{2}{*}{Outdoor City}
& SANA-WM
& 12.79
& 0.2479
& 0.5767
& 7.1726
& 1.7286
& 1.7495 \\
& \methodname
& \textbf{13.05}
& \textbf{0.2676}
& \textbf{0.5644}
& \textbf{6.6284}
& \textbf{1.6204}
& \textbf{1.6402} \\

\multirow{2}{*}{Outdoor Nature}
& SANA-WM
& 12.97
& 0.2776
& 0.5849
& 6.9223
& 2.9796
& 2.9997 \\
& \methodname
& \textbf{13.94}
& \textbf{0.2960}
& \textbf{0.5456}
& \textbf{5.8709}
& \textbf{2.9053}
& \textbf{2.9150} \\

\midrule
\multicolumn{8}{l}{\textit{Hard-Trajectory Split}} \\
\cmidrule(lr){1-8}

\multirow{2}{*}{Game Style}
& SANA-WM
& 13.75
& 0.3443
& 0.5140
& \textbf{8.1256}
& \textbf{1.4874}
& \textbf{1.5398} \\
& \methodname
& \textbf{14.55}
& \textbf{0.3797}
& \textbf{0.4956}
& 9.1194
& 1.5393
& 1.5961 \\

\multirow{2}{*}{Indoor}
& SANA-WM
& 12.96
& 0.3148
& 0.5846
& 22.9948
& 0.9594
& 1.2115 \\
& \methodname
& \textbf{12.98}
& \textbf{0.3248}
& \textbf{0.5774}
& \textbf{14.8387}
& \textbf{0.7795}
& \textbf{0.9455} \\

\multirow{2}{*}{Outdoor City}
& SANA-WM
& 12.46
& 0.2631
& 0.5644
& 13.5222
& 1.6371
& 1.7186 \\
& \methodname
& \textbf{12.89}
& \textbf{0.2822}
& \textbf{0.5585}
& \textbf{11.4256}
& \textbf{1.5799}
& \textbf{1.6459} \\

\multirow{2}{*}{Outdoor Nature}
& SANA-WM
& 13.03
& 0.2667
& 0.5242
& 8.0424
& 2.5758
& 2.6078 \\
& \methodname
& \textbf{13.61}
& \textbf{0.2702}
& \textbf{0.5220}
& \textbf{6.7236}
& \textbf{2.2810}
& \textbf{2.3046} \\

\bottomrule
\end{tabular}
}
\end{table*}

Overall, the category-level results show that the improvements are not concentrated in a single scene type.
Stage~1 provides the most consistent gains, while downstream refinement introduces several category-specific trade-offs.
At the aggregate level, \methodname~improves all three revisit-consistency metrics under each evaluated inference mode.

\subsubsection{MBench-A Subset Results}
\label{app:mbench_subsets}

MBench-A contains $547$ samples across four official subsets: Causal, Human, Environment, and Object.
For each sample, we first average its matched revisit pairs and then compute the macro average across samples.
Table~\ref{tab:mbench_subset_results} reports the complete subset-level results under the three evaluated inference modes.
Across all $12$ mode--subset combinations, \methodname~simultaneously improves PSNR, SSIM, and LPIPS, demonstrating consistent generalization across diverse scene and interaction types.

\begin{table*}[t]
\centering
\caption{
Detailed Top-$1$ results on the four official MBench-A subsets under different inference modes.
Best results within each subset and mode are shown in bold.
}
\label{tab:mbench_subset_results}
\resizebox{0.8\textwidth}{!}{
\begin{tabular}{lclccc}
\toprule
Subset & $n$ & Method
& PSNR $\uparrow$
& SSIM $\uparrow$
& LPIPS $\downarrow$ \\
\midrule

\multicolumn{6}{l}{\textit{Stage~1}} \\
\cmidrule(lr){1-6}

\multirow{2}{*}{Causal} & \multirow{2}{*}{100}
& SANA-WM
& 10.4068 & 0.28982 & 0.65461 \\
& & \methodname
& \textbf{11.0022} & \textbf{0.31830} & \textbf{0.61659} \\

\multirow{2}{*}{Human} & \multirow{2}{*}{120}
& SANA-WM
& 9.5147 & 0.26845 & 0.67977 \\
& & \methodname
& \textbf{10.2161} & \textbf{0.30275} & \textbf{0.63759} \\

\multirow{2}{*}{Environment} & \multirow{2}{*}{227}
& SANA-WM
& 10.4140 & 0.24056 & 0.59436 \\
& & \methodname
& \textbf{11.4626} & \textbf{0.27144} & \textbf{0.54857} \\

\multirow{2}{*}{Object} & \multirow{2}{*}{100}
& SANA-WM
& 10.1755 & 0.23386 & 0.59953 \\
& & \methodname
& \textbf{10.9412} & \textbf{0.25989} & \textbf{0.55808} \\

\midrule
\multicolumn{6}{l}{\textit{Causal AR Refine}} \\
\cmidrule(lr){1-6}

\multirow{2}{*}{Causal} & \multirow{2}{*}{100}
& SANA-WM
& 12.4358 & 0.33422 & 0.67586 \\
& & \methodname
& \textbf{12.6230} & \textbf{0.34587} & \textbf{0.66550} \\

\multirow{2}{*}{Human} & \multirow{2}{*}{120}
& SANA-WM
& 11.8551 & 0.30453 & 0.70110 \\
& & \methodname
& \textbf{12.2041} & \textbf{0.31959} & \textbf{0.68707} \\

\multirow{2}{*}{Environment} & \multirow{2}{*}{227}
& SANA-WM
& 12.6355 & 0.29205 & 0.62765 \\
& & \methodname
& \textbf{13.0913} & \textbf{0.30450} & \textbf{0.61762} \\

\multirow{2}{*}{Object} & \multirow{2}{*}{100}
& SANA-WM
& 12.2440 & 0.27644 & 0.64322 \\
& & \methodname
& \textbf{12.6283} & \textbf{0.28724} & \textbf{0.63458} \\

\midrule
\multicolumn{6}{l}{\textit{Bidirectional Refine}} \\
\cmidrule(lr){1-6}

\multirow{2}{*}{Causal} & \multirow{2}{*}{100}
& SANA-WM
& 12.8197 & 0.36602 & 0.58537 \\
& & \methodname
& \textbf{13.2118} & \textbf{0.37457} & \textbf{0.57395} \\

\multirow{2}{*}{Human} & \multirow{2}{*}{120}
& SANA-WM
& 12.0657 & 0.33279 & 0.62407 \\
& & \methodname
& \textbf{12.5213} & \textbf{0.34830} & \textbf{0.60919} \\

\multirow{2}{*}{Environment} & \multirow{2}{*}{227}
& SANA-WM
& 13.3562 & 0.32284 & 0.52598 \\
& & \methodname
& \textbf{14.1215} & \textbf{0.33725} & \textbf{0.51069} \\

\multirow{2}{*}{Object} & \multirow{2}{*}{100}
& SANA-WM
& 13.1585 & 0.31224 & 0.53320 \\
& & \methodname
& \textbf{13.6448} & \textbf{0.31892} & \textbf{0.52774} \\

\bottomrule
\end{tabular}
}
\end{table*}

The subset-level results confirm that the aggregate improvements are not driven by a particular subset.
Although the gain magnitude varies across subsets and refinement modes, \methodname~improves all three revisit-consistency metrics in every evaluated MBench-A setting.

\subsection{Additional Qualitative Results}
\label{app:additional_visuals}

\subsubsection{Stage~1}
\label{app:stage1_visuals}

\begin{figure*}[p]
    \centering

    \includegraphics[width=\textwidth]
    {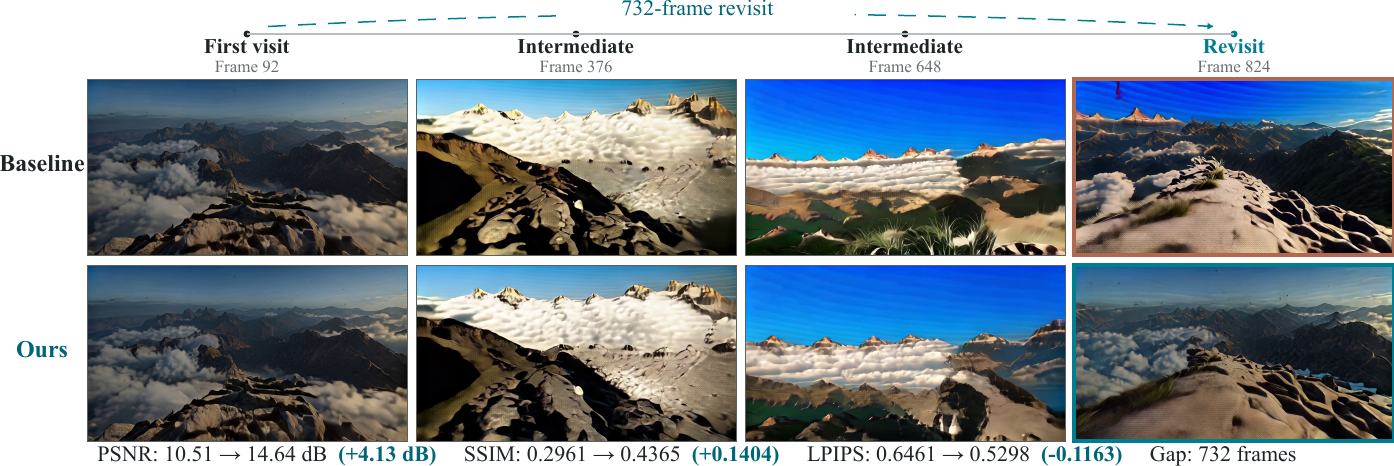}

    \vspace{0.8mm}

    \includegraphics[width=\textwidth]
    {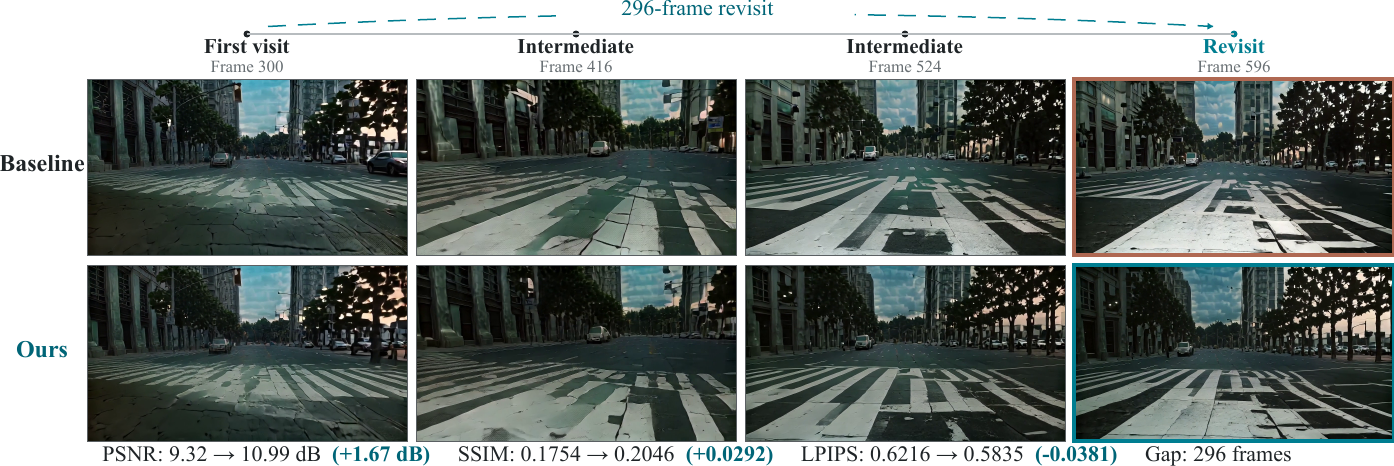}

    \vspace{0.8mm}

    \includegraphics[width=\textwidth]
    {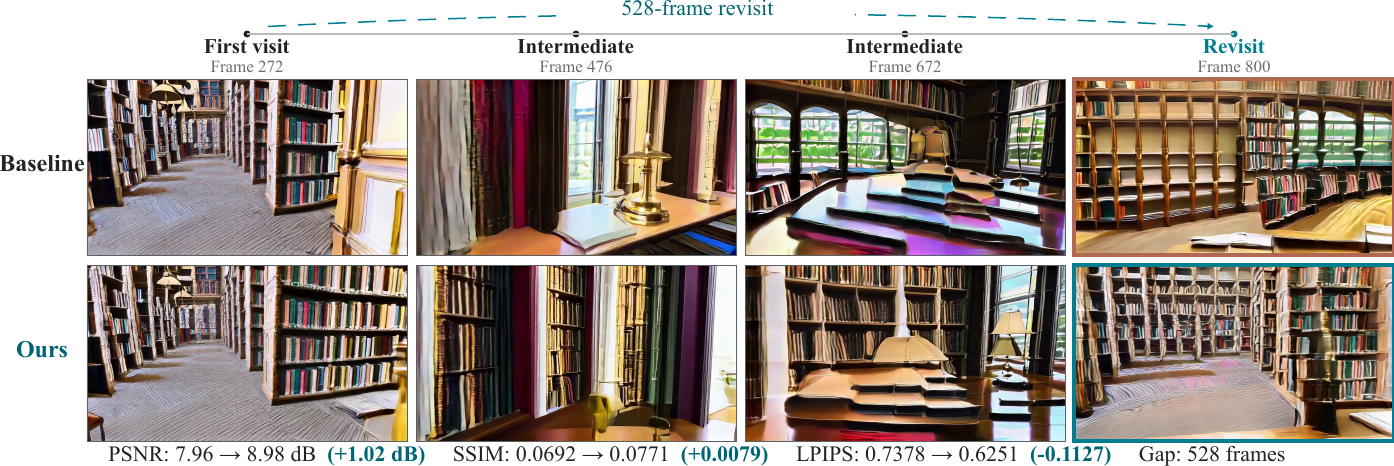}

    \vspace{0.8mm}

    \includegraphics[width=\textwidth]
    {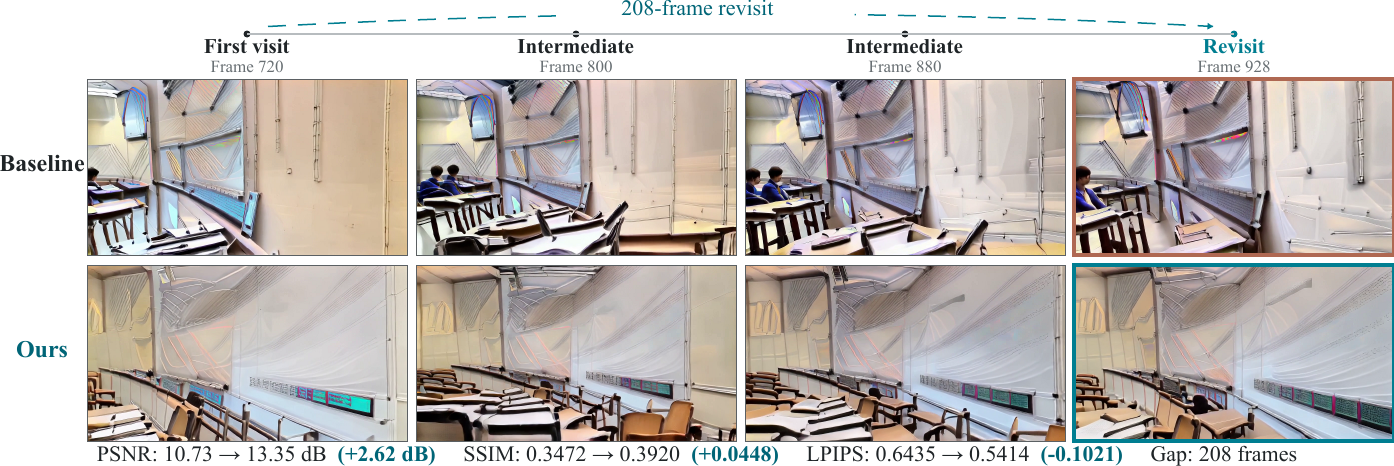}

    \caption{
    Additional qualitative comparisons for Stage~1 outputs.
    Each example shows the first visit, two intermediate views, and a later revisit along the same camera trajectory.
    Revisit consistency is evaluated between the first-visit and revisit frames within each method.
    }
    \label{fig:stage1_visuals}
\end{figure*}

Figure~\ref{fig:stage1_visuals} provides additional qualitative comparisons for the Stage~1 outputs before downstream refinement.
In the first example, the baseline exhibits clear drift in the revisit view, with noticeable changes in the mountain contours, foreground ridge shape, and cloud distribution. In contrast, \methodname~better preserves the overall mountain silhouette, the foreground terrain layout, and the spatial relationship between the ridge and the cloud layer.
In the second example, the baseline alters the street geometry upon revisit, leading to inconsistent crosswalk patterns, road markings, and building alignment. \methodname~more faithfully maintains the width and orientation of the crosswalk, the road layout, and the arrangement of nearby buildings and trees.
In the third example, the baseline largely loses the original library-like structure and replaces it with a substantially different interior configuration. \methodname~better preserves the bookshelf arrangement, corridor layout, desk region, and window placement, resulting in a revisit view that remains much closer to the first visit.
In the fourth example, the baseline shows substantial changes in the classroom structure, including distortions in the wall geometry, seating layout, and front display region. By comparison, \methodname~better maintains the overall room geometry, the arrangement of desks and chairs, and the long horizontal display panels along the wall.
Overall, these examples show that \methodname~improves long-range scene consistency already at Stage~1, before any subsequent refinement.

\subsubsection{Stage~1 + AR Refine}
\label{app:ar_refine_visuals}

\begin{figure*}[p]
    \centering

    \includegraphics[width=\textwidth]
    {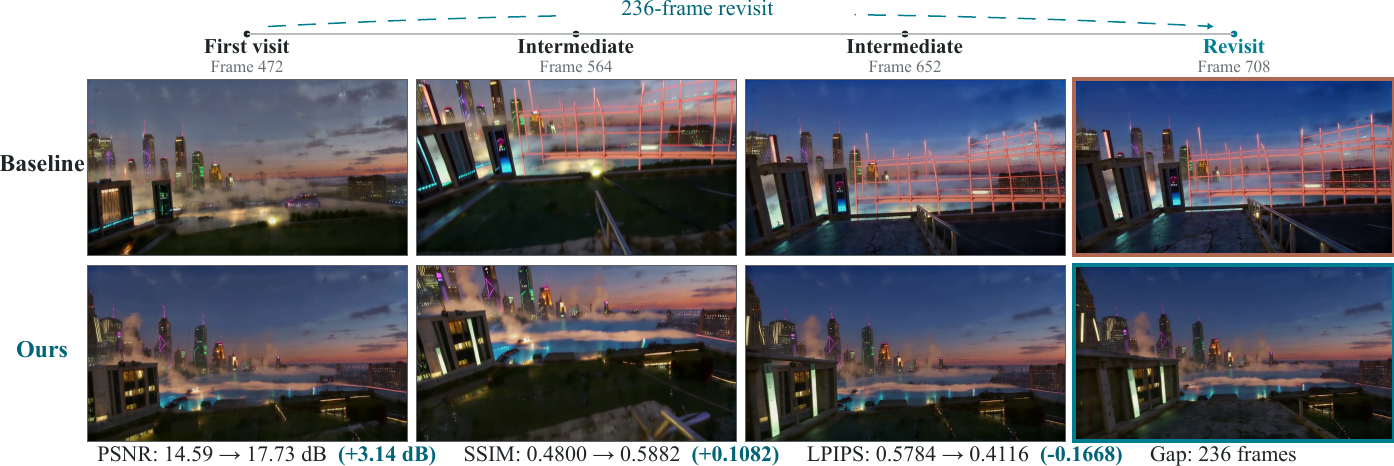}

    \vspace{0.8mm}

    \includegraphics[width=\textwidth]
    {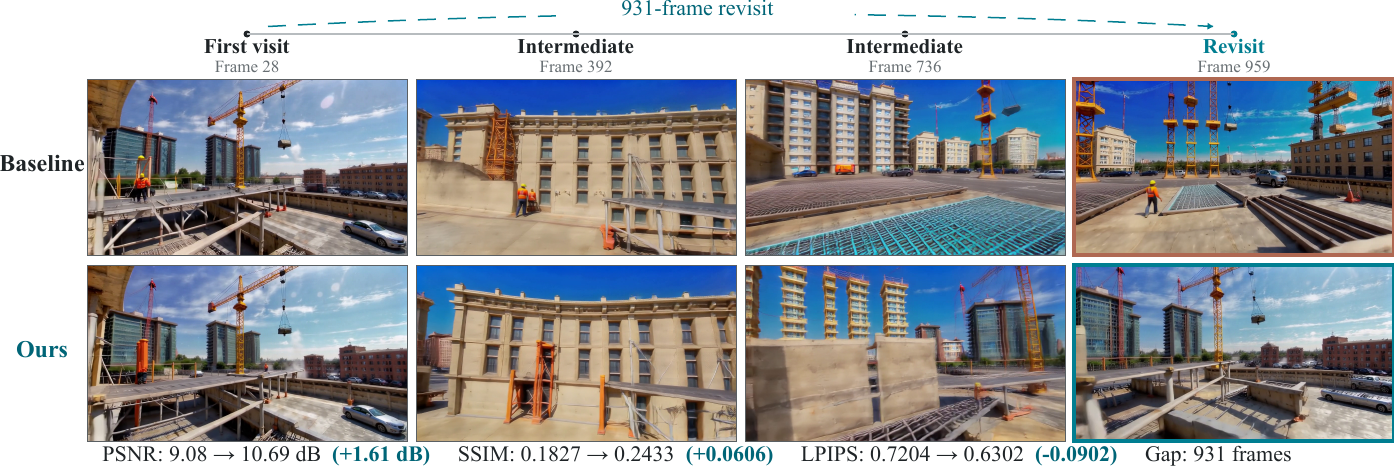}

    \vspace{0.8mm}

    \includegraphics[width=\textwidth]
    {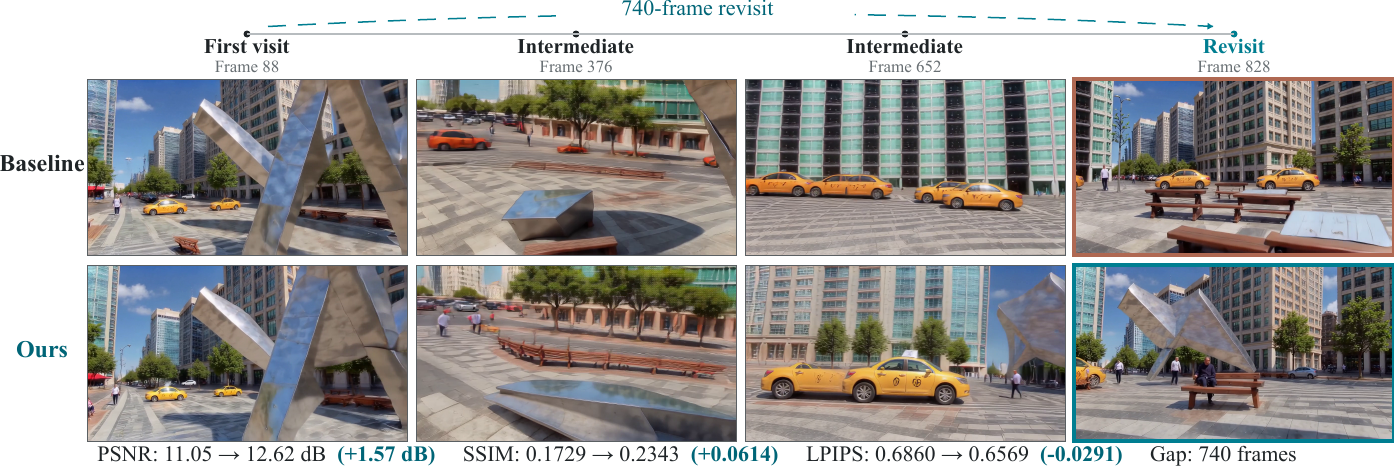}

    \vspace{0.8mm}

    \includegraphics[width=\textwidth]
    {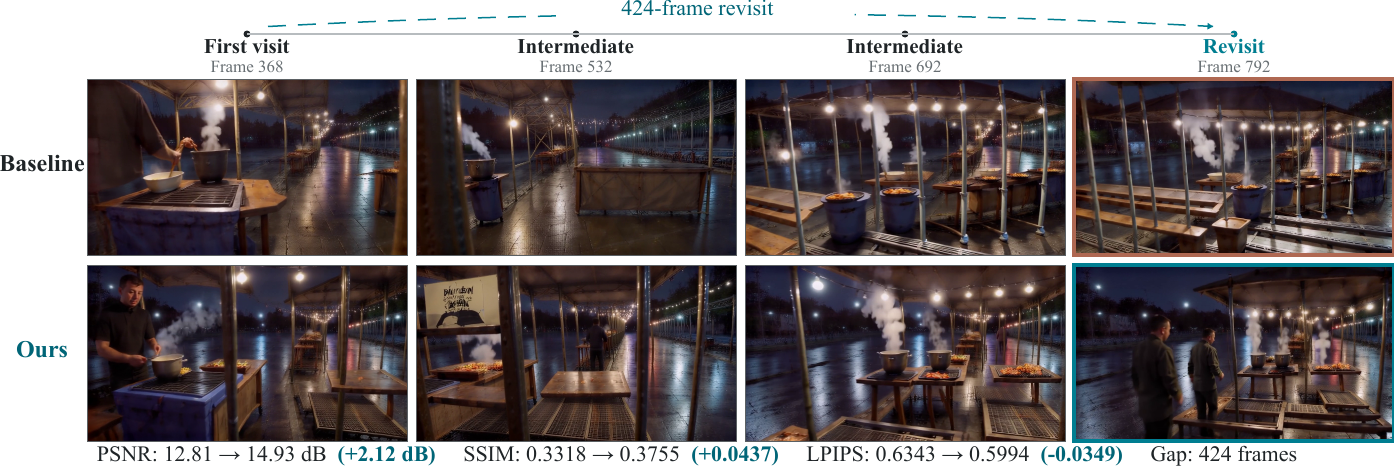}

    \caption{
    Additional qualitative comparisons under causal AR refinement.
    Each example shows the first visit, two intermediate views, and a later revisit along the same camera trajectory.
    Revisit consistency is evaluated between the first-visit and revisit frames within each method.
    }
    \label{fig:ar_refine_visuals}
\end{figure*}

Figure~\ref{fig:ar_refine_visuals} provides additional qualitative comparisons under causal AR refinement.
Across all examples, \methodname~better preserves previously observed scene content when the camera revisits the same region after an extended rollout.
In the first example, the baseline substantially changes the distant skyline and foreground layout upon revisit, whereas \methodname~better preserves the overall scene structure and characteristic background content.
In the second example, the baseline noticeably alters the foreground geometry and loses much of the original structural layout, while \methodname~better maintains the major scene elements and their spatial arrangement.
In the third example, the baseline largely loses the distinctive foreground structure and changes the arrangement of surrounding objects, whereas \methodname~more faithfully preserves these characteristic elements and their relative geometry.
In the fourth example, \methodname~better maintains the spatial arrangement of tables, people, and illuminated structures, while the baseline exhibits substantial changes in the revisited scene.
Overall, these examples demonstrate improved preservation of both global scene structure and distinctive visual content over long-range revisits.

\subsubsection{Stage~1 + Bidirectional Refine}
\label{app:bi_refine_visuals}

\begin{figure*}[p]
    \centering

    \includegraphics[width=\textwidth]
    {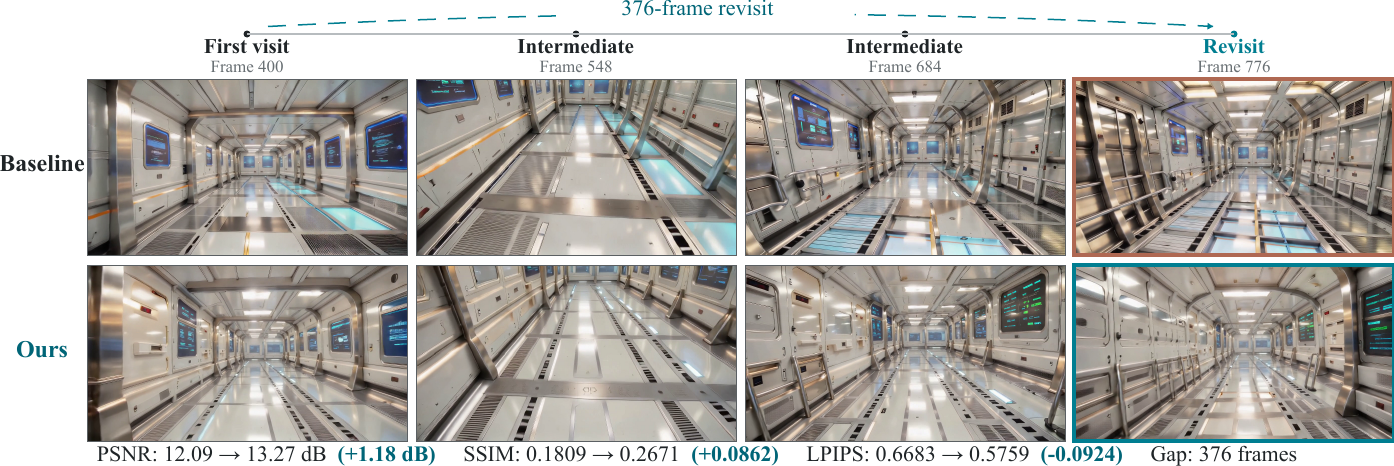}

    \vspace{0.8mm}

    \includegraphics[width=\textwidth]
    {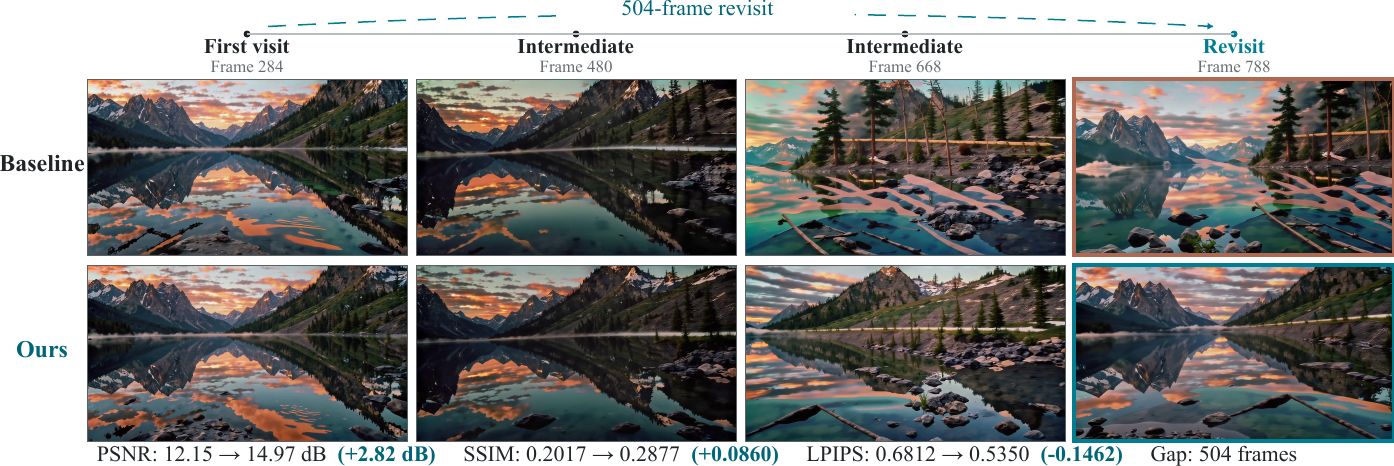}

    \vspace{0.8mm}

    \includegraphics[width=\textwidth]
    {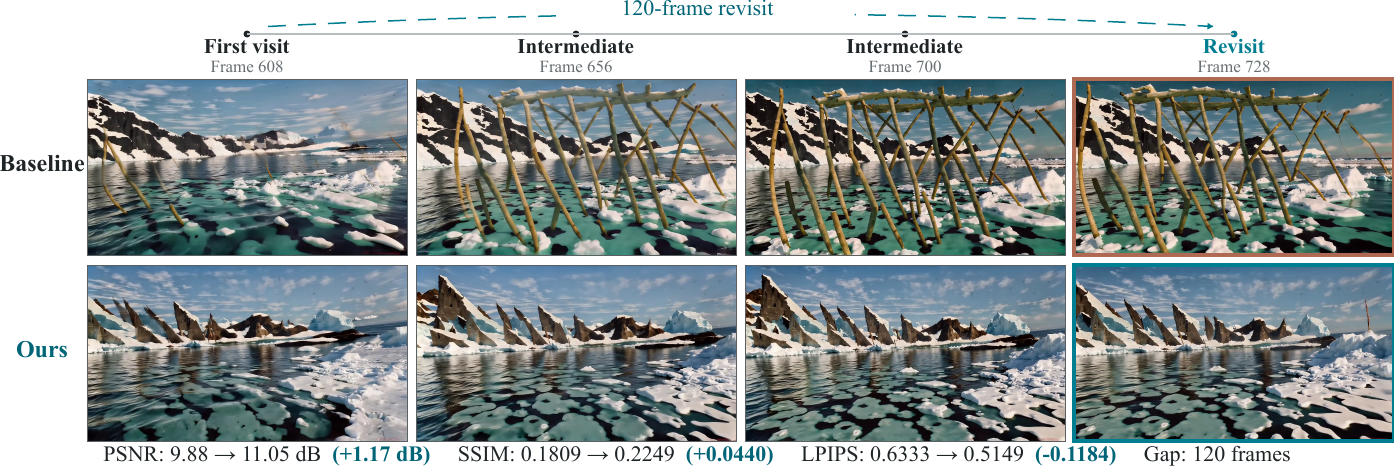}

    \vspace{0.8mm}

    \includegraphics[width=\textwidth]
    {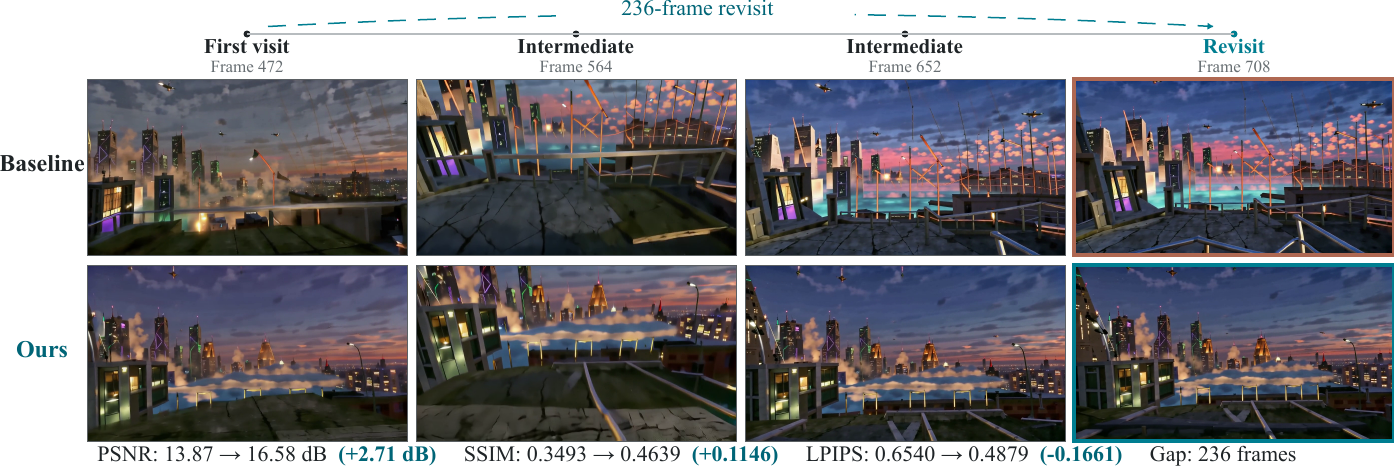}

    \caption{
    Additional qualitative comparisons under bidirectional refinement.
    Each example shows the first visit, two intermediate views, and a later revisit along the same camera trajectory.
    Revisit consistency is evaluated between the first-visit and revisit frames within each method.
    }
    \label{fig:bi_refine_visuals}
\end{figure*}

Figure~\ref{fig:bi_refine_visuals} provides additional qualitative comparisons under bidirectional refinement.
In the first example, the baseline exhibits noticeable drift in the corridor geometry upon revisit, including changes in the floor pattern, wall modules, and spatial arrangement of the side structures. In contrast, \methodname~better preserves the corridor layout, perspective, and characteristic interior elements.
In the second example, the baseline substantially alters the lake surface, shoreline, and surrounding landscape, with pronounced distortions appearing in the revisited view. \methodname~more faithfully maintains the mountain silhouette, tree line, shoreline geometry, and their reflections on the water.
In the third example, the baseline introduces a large wooden structure that is absent from the first visit and significantly changes the foreground composition. \methodname~avoids this spurious structural change and better preserves the open-water layout, ice distribution, and distant mountain ridge.
In the fourth example, the baseline shows considerable changes in the rooftop geometry, foreground structures, and distant skyline, whereas \methodname~better retains their relative arrangement and the overall scene identity.
Together, these examples further demonstrate that selective long-range memory retrieval improves the preservation of both global geometry and distinctive scene content during bidirectional refinement.

\end{document}